\documentclass[sigconf]{acmart}
\usepackage{xspace}
\usepackage{booktabs}
\usepackage{graphicx}
\usepackage{subfigure}
\usepackage{enumitem}

\AtBeginDocument{%
  }

\setcopyright{acmcopyright}
\copyrightyear{2023}
\acmYear{2023}
\acmDOI{10.1145/3589132.3625644}

\acmConference[SIGSPATIAL ’23]{The 31th International Conference on Advances in Geographic Information Systems}{November 13-16, 2023}{Hamburg, Germany}
\acmPrice{15.00}
\acmISBN{979-8-4007-0168-9/23/11}

\acmSubmissionID{8433}

\begin{document}
% \newcommand{\projtitle}{STGTrans\xspace}
% \newcommand{\projtitle}{MobGT\xspace}

% \title{Spatial-Temporal Graph Transformer for Next Point-of-Interest Recommendation}
\title{Revisiting Mobility Modeling with Graph: A Graph Transformer Model for Next Point-of-Interest Recommendation}

\author{Xiaohang Xu}
\affiliation{%
  \institution{The University of Tokyo}
  % \streetaddress{P.O. Box 1212}
  \city{Tokyo}
  % \state{Tokyo}
  \country{Japan}
}
\email{xhxu@g.ecc.u-tokyo.ac.jp}

\author{Toyotaro Suzumura}
\affiliation{%
  \institution{The University of Tokyo}
  % \streetaddress{P.O. Box 1212}
  \city{Tokyo}
  % \state{Tokyo}
  \country{Japan}
}
\email{suzumura@ds.itc.u-tokyo.ac.jp}

\author{Jiawei Yong}
\affiliation{%
  \institution{Toyota Motor Corporation}
  % \streetaddress{P.O. Box 1212}
  \city{Tokyo}
  % \state{Tokyo}
  \country{Japan}
}
\email{jiawei_yong@mail.toyota.co.jp}

\author{Masatoshi Hanai}
\affiliation{%
  \institution{The University of Tokyo}
  % \streetaddress{P.O. Box 1212}
  \city{Tokyo}
  % \state{Tokyo}
  \country{Japan}
}
\email{hanai@ds.itc.u-tokyo.ac.jp}

\author{Chuang Yang}
\affiliation{%
  \institution{The University of Tokyo}
  % \streetaddress{P.O. Box 1212}
  \city{Tokyo}
  % \state{Tokyo}
  \country{Japan}
}
\email{chuang.yang@csis.u-tokyo.ac.jp}

\author{Hiroki Kanezashi}
\affiliation{%
  \institution{The University of Tokyo}
  % \streetaddress{P.O. Box 1212}
  \city{Tokyo}
  % \state{Tokyo}
  \country{Japan}
}
\email{hkanezashi@ds.itc.u-tokyo.ac.jp}

\author{Renhe Jiang}
\affiliation{%
  \institution{The University of Tokyo}
  % \streetaddress{P.O. Box 1212}
  \city{Tokyo}
  % \state{Tokyo}
  \country{Japan}
}
\email{jiangrh@csis.u-tokyo.ac.jp}

\author{Shintaro Fukushima}
\affiliation{%
  \institution{Toyota Motor Corporation}
  % \streetaddress{P.O. Box 1212}
  \city{Tokyo}
  % \state{Tokyo}
  \country{Japan}
}
\email{s_fukushima@mail.toyota.co.jp}

\renewcommand{\shortauthors}{X. Xu et al.}

\begin{abstract}
Next Point-of-Interest (POI) recommendation plays a crucial role in urban mobility applications. 
Recently, POI recommendation models based on Graph Neural Networks (GNN) have been extensively studied and achieved, however, the effective incorporation of both spatial and temporal information into such GNN-based models remains challenging.
% Recently, POI recommendation models based on Graph Neural Networks (GNN) have been extensively studied and achieved high accuracy by leveraging graphs to globally capture intricate user-location relations.
% However, the effective incorporation of both spatial and temporal information into such GNN-based models remains challenging.
Temporal information is extracted from users' trajectories, while spatial information is obtained from POIs.
Extracting distinct fine-grained features unique to each piece of information is difficult since temporal information often includes spatial information, as users tend to visit nearby POIs.
To address the challenge, we propose \textbf{\underline{Mob}}ility \textbf{\underline{G}}raph \textbf{\underline{T}}ransformer (MobGT) that enables us to fully leverage graphs to capture both the spatial and temporal features in users' mobility patterns.
MobGT combines individual spatial and temporal graph encoders to capture unique features and global user-location relations. 
Additionally, it incorporates a mobility encoder based on Graph Transformer to extract higher-order information between POIs. 
To address the long-tailed problem in spatial-temporal data, MobGT introduces a novel loss function, Tail Loss.
% \projtitle involves three key techniques. % : (i) Graph Transformer-based model, (ii) Individual spatial and temporal graph encoder, and (iii) Gradient TailLoss.
% First, we design individual spatial and temporal graph encoders for capturing their corresponding unique features and the global user-location relations at the same time.
% Second, we design a trajectory encoder based on the state-of-the-art Graph Transformer-based model that can extract higher-order information between different POIs.
% Although these two techniques can enhance the ability to capture fine-grained features, they may suffer from performance degradation due to the long-tailed problem in the locational data, i.e., the majority of locations are visited a few times.
% Thus, finally, we introduce a novel loss function, called Tail Loss, to solve the long-tailed problem.
Experimental results demonstrate that MobGT outperforms state-of-the-art models on various datasets and metrics, achieving 24\% improvement on average. 
Our codes are available at \url{https://github.com/Yukayo/MobGT}.

\end{abstract}

\begin{CCSXML}
<ccs2012>
<concept>
<concept_id>10002951.10003227.10003236.10003101</concept_id>
<concept_desc>Information systems~Location based services</concept_desc>
<concept_significance>500</concept_significance>
</concept>
</ccs2012>

<ccs2012>
   <concept>
       <concept_id>10002951.10003317.10003347.10003350</concept_id>
       <concept_desc>Information systems~Recommender systems</concept_desc>
       <concept_significance>500</concept_significance>
       </concept>
 </ccs2012>

\end{CCSXML}

\ccsdesc[500]{Information systems~Location based services}
\ccsdesc[500]{Information systems~Recommender systems}
\keywords{Point-of-Interest (POI), Transformer, Graph Neural Network, Human Mobility, Recommender System}

% \received{2 February 2013}

\maketitle

\section{Introduction}
\label{sec:1_intro}
With the proliferation of location-based social networks and mobile applications, a large amount of check-in information with spatial-temporal labels has been generated. 
Moreover, with the widespread use of location-based services in vehicles, such as GPS systems and ride-sharing applications, people also leverage these spatial-temporal data for mobility applications.
Point-of-interest (POI) recommendation is a hot downstream task based on the spatial-temporal data that has been garnering more attention~\cite{zhao2020go,han2021point,rao2022graph}.

Traditional POI recommendation studies have primarily focused on extracting local information from individual users' trajectories~\cite{wu2019long,liu2022real,wu2020personalized,feng2018deepmove,sun2020go,jiang2022will,chen2020dualsin}.
However, recent studies have demonstrated significant performance improvements by also extracting global information from user-location relationships~\cite{yang2019revisiting,rao2022graph,lim2020stp,xue2021mobtcast}.
The former approaches are only based on sequential models, such as recurrent neural networks (RNNs) and Transformer, that treat users' trajectories simply as sequences.
The latter approaches are based on the integration of sequential models and graph models.
In addition to the capturing of local information via sequential models, such graph-based approaches also extract the global relations between users and locations through all the historical trajectories utilizing graph neural networks (GNNs).
In this paper, we focus on the graph-based approach.
Although the previous graph-based models have achieved significant results in the next POI recommendation, they still suffer from limitations that may result in poor recommendation performance.
% the model's inability to achieve high-precision recommendation performance.

% \begin{figure}[!t]
% \centering	
% \includegraphics[width=0.5\textwidth]{fig/Trajectory.pdf}
% \caption{Two users' trajectories exhibit similar visited POI pairs, and a loop and similar POI pairs with different distances exist in the trajectory of user A.}
% % \vspace{-0.5cm}
% \label{fig:trajectory}
% \end{figure}

\begin{figure}[!t]
\centering	
\includegraphics[width=0.48\textwidth]{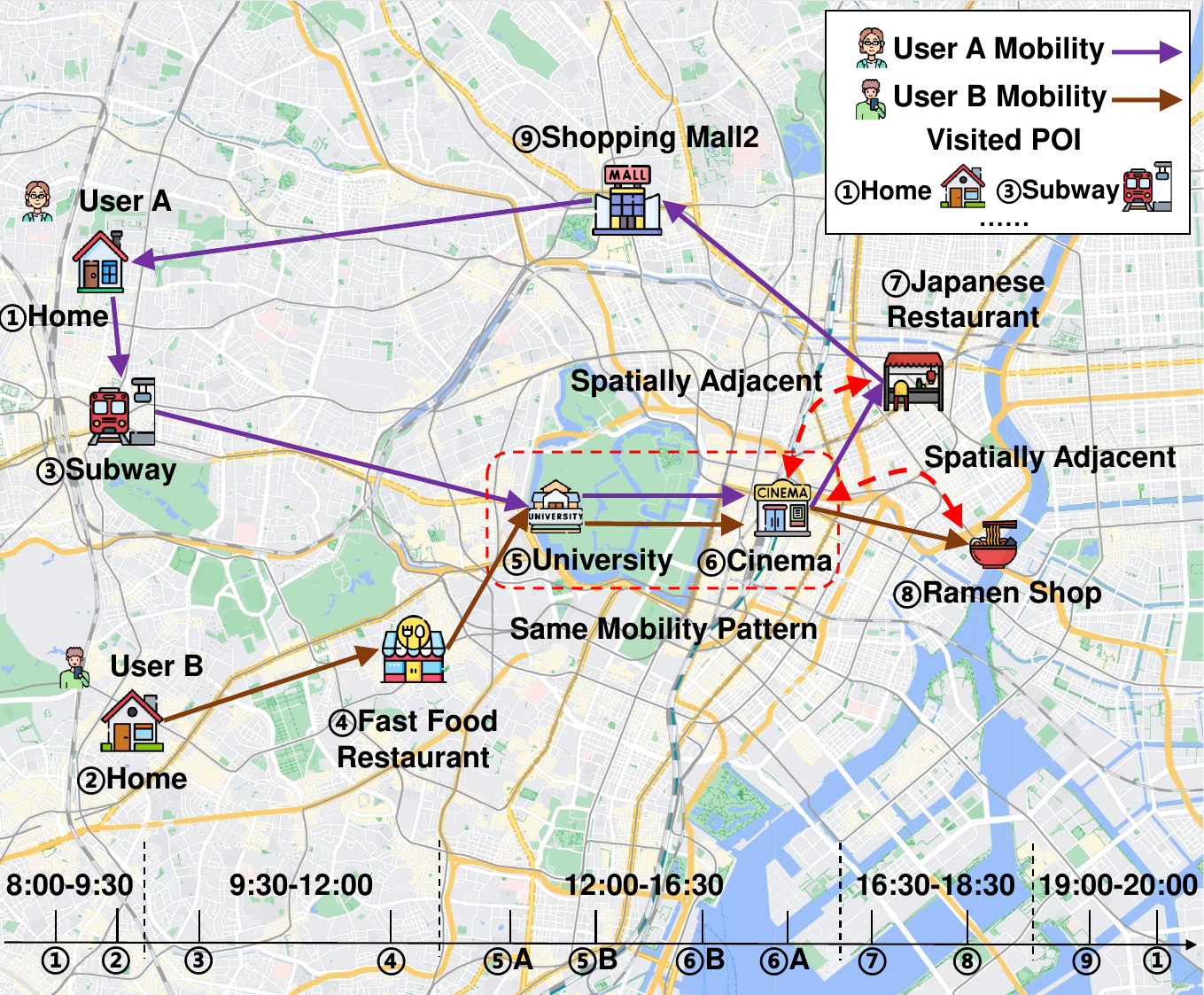}
\caption{Two users’ trajectories exhibit the same mobility pattern (from university to cinema) in different timestamps, and some POI pairs with the same category (from cinema (entertainment) to the Japanese restaurant or Ramen shop (food)) exist in spatially adjacent relationships. (Icons were designed by Flaticon)}
\vspace{-0.4cm}
\label{fig:trajectory}
\end{figure}

Firstly and most significantly, existing works on graph-based spatial-temporal modeling lack effective integration of both spatial and temporal information from a global view~\cite{li2022hmgcl, jiang2021transfer}.
% For example, Rao et al.~\cite{rao2022graph} proposed the STKG model to integrate spatial-temporal information for predicting the next POI. 
% While the aim of the model was to incorporate both spatial-temporal aspects using a Knowledge Graph, only temporal information was utilized in its final implementation.
% They argue that temporal relations have already implicitly captured users' geographical visiting property since people tend to visit places close to themselves. 
The recent state-of-the-art models consider incorporating both spatial-temporal features using Knowledge Graph~\cite{rao2022graph}.
But, they argue that temporal relations have already implicitly captured users' spatial visiting property and implemented their model without the spatial modeling part since people tend to visit places close to themselves. 
% We oppose this argument as a user's next visit is not only determined by temporal context, but also by spatial context if using transportation modes such as cars. 
However, the user's subsequent location is influenced not only by temporal context but also by spatial context.
% For instance, in Fig.~\ref{fig:trajectory}, both users A and B depart from a university and visit different restaurants before visiting the same shopping mall, indicating a similar transition pattern.
For example, in Fig.~\ref{fig:trajectory}, the cinema and two distinct types of restaurants are spatially adjacent, indicating that we can recommend the Japanese restaurant visited by user A to user B after watching a movie by utilizing spatially adjacent. 
% However, if we only rely on temporal information, we cannot explicitly obtain the existence of this spatial adjacency between the two restaurants.
It's challenging for the existing graph-based model to capture such spatial adjacency with only temporal information. 
% How can the model capture such mobility patterns with only temporal information?
% For example, in Fig~\ref{fig:trajectory}, both user A and user B visit the same business district, with user A check-in a Ramen shop and user B check-in a Japanese restaurant. 
% How can we use spatial information to capture such similar food preferences and mobility patterns?
% As for Traj1 of user A and Traj2 of user B, they both departed from a university and went to different restaurants before visiting the same shopping mall. 
% The transition pattern is very similar.
% How can the model capture such mobility patterns?
% Therefore, we propose MobGT to effectively integrate both spatial and temporal information to learn fine-grained user mobility patterns.

% Second, capturing more complex and fine-grained user behaviors from individual(local) trajectories is still challenging even though the existing work typically has been focused on.
Secondly, capturing more intricate and nuanced user behaviors from individual (local) mobility remains challenging.
In the previous graph-based POI recommendation models~\cite{rao2022graph,yang2022getnext}, local trajectories are encoded through sequential models even though the global information is encoded by GNNs.
However, such sequential modeling is limited in handling high-order spatial relationships, for instance, returning to the starting point after visiting multiple locations or repeating similar mobility patterns under different times and contexts.
In Fig.~\ref{fig:trajectory}, user A departs from home, visits several distinct locations, and eventually returns home in the evening, forming a closed-loop trajectory. 
Simultaneously, both user A and user B move from the university to the cinema, exhibiting repeated mobility patterns. 
It is hard for the sequential model to capture such cyclic mobility patterns and the same mobility pattern in the trajectory.

\iffalse
Such sequential modeling would have a limit in utilizing higher-order location-location relationships in individuals' trajectories.
% For example, in Fig.~\ref{fig:trajectory}, user B visited three consecutive "restaurant-shopping mall" pairs with varying distances between different pairs and returned to the university afterward.
For example, in Fig.~\ref{fig:trajectory}, user A starts from her home and returns home in the evening, forming a closed-loop trajectory.
In addition, both user A and user B visit the cinema from the university, resulting in them having the same mobility pattern.
It is hard for the sequential model to capture such cyclic mobility patterns and the same mobility pattern in the trajectory. 
\fi

Finally, as data obtained from POI check-in services inherently involve long-tailed problems, that is, most locations are visited a few times, the development of effective training methods that can utilize low-frequent location data is as significant as the design of the model itself to achieve fine-grained prediction. 

To address these three challenges, we propose a novel \textbf{\underline{Mob}}ility \textbf{\underline{G}}raph \textbf{\underline{T}}ransformer (MobGT) model.  
Our key contributions are summarized as follows:
\begin{itemize}[leftmargin=*]
\item We propose global mobility GNN modules from three perspectives: spatial, temporal, and POI categories, to model the user's POI transform pattern cooperatively.

\item We propose a graph-based Transformer model to model the local mobility graph of each user, which can utilize explicit contextual information to enrich the representation results and use a specialized encoding module to capture high-order dependency relationships in the graph.

\item We propose a novel loss function, Tail Loss, to address the long-tailed problem in locational data.

\item We conduct extensive experiments on two publicly available real-world check-in datasets and a private vehicle GPS trajectory dataset. 
Our experimental results demonstrate that our proposed MobGT outperforms the existing state-of-the-art models, achieving an average improvement of 24\%.
\end{itemize}
\section{Related Work}
\label{sec:2_relatedwork}

\subsection{Next POI Recommendation}
\label{sec:2_nextpoi}
In contrast to the conventional POI recommendation approach, which bears resemblance to product or news recommendation, the task of the next POI recommendation involves analyzing and identifying the most probable POI that a user will visit next, based on their recent or historical mobility pattern~\cite{yao2017serm}.
Early work more focused on non-deep learning-based recommendation approaches such as Markov Chain (MC)~\cite{cheng2013you,ye2013s,gambs2012next,liu2013personalized}, Collaborative Filtering (CF)~\cite{ye2010location,gao2018personalized}, and Matrix Factorization (MF)~\cite{liu2014exploiting,lian2014geomf,liu2013personalized,lian2018geomf++}.
For example, ~\cite{liu2013personalized} extended the POI recommendation that combines both CF and MF.
Firstly, it utilized CF to learn the current user's POI transition pattern by studying the POI transition patterns of other similar users. 
Then, it clustered similar users and employed MF to predict the user's interested POIs.
Similarly, Zhang et al.~\cite{zhang2016point} extended the application of additive Markov chain~\cite{zhang2015spatiotemporal} by using categorical, geographic, and other contextual information.
Meanwhile, some studies have also explored the use of Markov chains or Bayesian methods to provide personalized next POI recommendations~\cite{feng2015personalized,chang2018content,he2016inferring}.
However, these methods also have obvious deficiencies, they are unable to effectively learn overall trends of movement within sequences.
For example, Markov chains mainly rely on local information in the sequence but do not learn enough about the overall trend.

Deep learning models, particularly those based on RNN and LSTM, have been widely adopted to address the next POI recommendation problem~\cite{lian2020geography,liu2016predicting,yang2020location,zhao2020discovering}. 
The early representative work was that Liu et al. proposed the Spatial-Temporal Recurrent Neural Networks (ST-RNN)~\cite{liu2016predicting}, which extends the traditional RNN by incorporating the transformation matrix of temporal and spatial features.
In 2020, Yang et al. proposed the Flashback~\cite{yang2020location} model, which focuses primarily on sparse user information and effectively captures historical trajectory information through modifications to the return value of the RNN hidden state. 
Due to the limited ability of RNNs to capture users' long-term interests, many researchers have proposed POI recommendation methods based on LSTM. 
In HST-LSTM~\cite{kong2018hst}, the authors utilized a layered structure to capture the periodicity of the mobility to achieve POI recommendation with LSTM structure. 
Feng et al. proposed the DeepMove~\cite{feng2018deepmove} architecture, which employs two distinct LSTM architectures and attention mechanisms to capture users' long-term and short-term interests, respectively. 
LSTPM~\cite{sun2020go} improved upon the design of DeepMove~\cite{feng2018deepmove} by incorporating a Geo-dilated LSTM module into the short-term interest module, to learn the geographical distribution of POIs in short-term interests. 
STAN~\cite{luo2021stan} showed the significance of non-adjacent POIs for recommendation results, using two self-attention layers to learn the correlation of adjacent and non-adjacent POIs. 
However, these works all employ sequential models to address the next POI recommendation problem, neglecting to learn movement patterns and POI distributions from a global perspective. 
% and also ignoring high-order features that capture the short-term interests of each user.
% \fi

\subsection{Graph-based POI Recommendation}
\label{sec:2_graph_poi}
Recently, researchers have proposed various graph-based POI recommendation models, including GTAG~\cite{xie2016learning}, STP-UDGAT~\cite{lim2020stp}, SGRec~\cite{li2021discovering}, and GETNext~\cite{yang2022getnext}. 
GTAG~\cite{xie2016learning} innovatively introduced the trajectory session into the graph, forming many connected edges through user-session-multiple check-in POIs, resulting in a large graph. 
However, the introduction of session nodes may lead to some GNN methods being unable to handle such large-scale graphs. 
STP-UDGAT~\cite{lim2020stp} proposed to establish a global graph between various POIs from the view of spatial and temporal to introduce more contextual information. 
However, all global graphs are undirected graphs, and the edge weight is almost constant, which loses the order relationship between POI-POI and blurs the accurate time and space information. SGRec~\cite{li2021discovering} used Graph Attention Networks (GAT) to capture the similarity between trajectories from the perspective of collaborative signals~\cite{wang2019neural} to improve the representation of POIs but ignores the spatial-temporal information of POIs. 
GETNext~\cite{yang2022getnext} aggregated all check-in POIs from the perspective of time pair to form a global graph and used GNN to generate corresponding POI embeddings for downstream POI recommendation.
However, for each user's trajectory, GETNext still uses the Transformer~\cite{vaswani2017attention} architecture to predict the next POI, which cannot model the explicit correlations between POIs within each trajectory, such as the spatial distance and visit frequency between POIs.

In contrast, our proposed model first constructs a unified global graph using three different global perspectives, which can contain POI geospatial information, POI temporal transition patterns, and category transition frequencies among POIs. 
Besides, for each user's mobility data, we convert it into a local mobility graph and use a graph-based Transformer to learn high-order relationships between POIs.
To the best of our knowledge, our model is the first next POI recommendation model to multi-view encode POI information based on the global graph and individual mobility subgraphs.

\section{Preliminary}
\label{sec:3_preliminaries}
% In this section, we present the key definitions and concepts relevant to the POI recommendation task.
In this section, we present the key definitions relevant to our task.
Let $U = \{u_1, u_2, ..., u_{|U|}\}$ be a set of $|U|$ users and $P = \{p_1, p_2, ..., p_{|P|}\}$ be a set of $|P|$ check-in locations (POIs).
% such as well-known brand names like KFC or McDonald's. 
Each location $p \in P$ is represented as a tuple $p = \{freq, cat, lat, lon\}$, including the visit frequency, category, latitude, and longitude.
Furthermore, let $T = \{t_1, t_2, ..., t_{|T|}\}$ represent the check-in times of each user at each POI.
We now define several key concepts in our work:
\begin{definition}[Check-in]
A check-in behavior is represented as a tuple $k = \{u, p, t\}$, meaning that a user $u$ visited and checked-in location $p$ at timestamp $t$.
\end{definition}

\begin{definition}[User Check-in Trajectory]
Given a user $u$, the user check-in trajectory $C_u$ is a series of check-in points, i.e. $C_u = \{k_1, k_2, ..., k_{|C_u|}\}$.
\end{definition}

% \begin{definition}[User Check-in Mobility]
% Given a user $u$, the user check-in mobility $C_u$ is a series of check-in points, i.e. \ $C_u = \{k_1, k_2, ..., k_{|C_u|}\}$.
% \end{definition}

\begin{definition}[Trajectory]
We define a trajectory as a sequence $C_u^{t_r}$ = $\{k_1$, $k_2$, ..., $k_{|C_u^{t_r}|}\}$ of $C_u$ ordered by time, with a timestamp $t_r$. 
% Specifically, for all $1 < i \le |C_u^{t_r}|$, the timestamp $t_i$ in $k_i$ belongs to $t_r$. 
% It should be noted that for data processing purposes, the trajectory is split into a fixed time interval, such as one day.
\end{definition}

% \begin{definition}[Mobility]
% We define a Mobility as a sequence $C_u^{t_r}$ = $\{k_1$, $k_2$, ..., $k_{|C_u^{t_r}|}\}$ of $C_u$ ordered by time, with a time range $t_r$. 
% Specifically, for all $1 < i \le |C_u^{t_r}|$, the timestamp $t_i$ in $k_i$ belongs to $t_r$. 
% % It should be noted that for data processing purposes, the trajectory is split into a fixed time interval, such as one day.
% \end{definition}

\begin{definition}[Next POI Recommendation]
Our objective is to generate a list of the POIs that a user $u$ is most likely to visit next.
% given a short trajectory and all historical mobility data for all users.
More specific, given all users' historical mobility trajectories $C = \{C_{u_1}, C_{u_2}, ..., C_{u_N}\}$ and a short-term trajectory $C_{u_i}^{t_r} = \{k_1, k_2, ..., k_{t_r}\}$ of any user $u_i \in U$, where $1 \le i \le N, 1 \le {t_r} \le |C_{u_i}^{t_r}|$, we need to output the top-$n$ POIs $\{k_{t_{r} + 1}^1, k_{t_{r} + 1}^2, ..., k_{t_{r} + 1}^n\}$ that the user $u_i$ is most likely to visit next in a timestamp $t_{r}$.
\end{definition}

% \begin{definition}[Next POI Recommendation]
% Our objective is to generate a list of the POIs that a user $u$ is most likely to visit next.
% % given a short trajectory and all historical mobility data for all users.
% More specific, given all users' historical mobility $C = \{C_{u_1}, C_{u_2}, ..., C_{u_N}\}$ and a short-term mobility $C_{u_i}^j = \{k_1, k_2, ..., k_j\}$ of any user $u_i \in U$, where $1 \le i \le N, 1 \le j \le |C_{u_i}|$, we need to output the top-$n$ POIs $\{k_{j + 1}^1, k_{j + 1}^2, ..., k_{j + 1}^n\}$ that the user $u_i$ is most likely to visit next.
% \end{definition}

\section{Methodology}
\label{sec:4_approach}

\begin{figure*}[!ht]
\centering	
\includegraphics[width=0.98\textwidth]{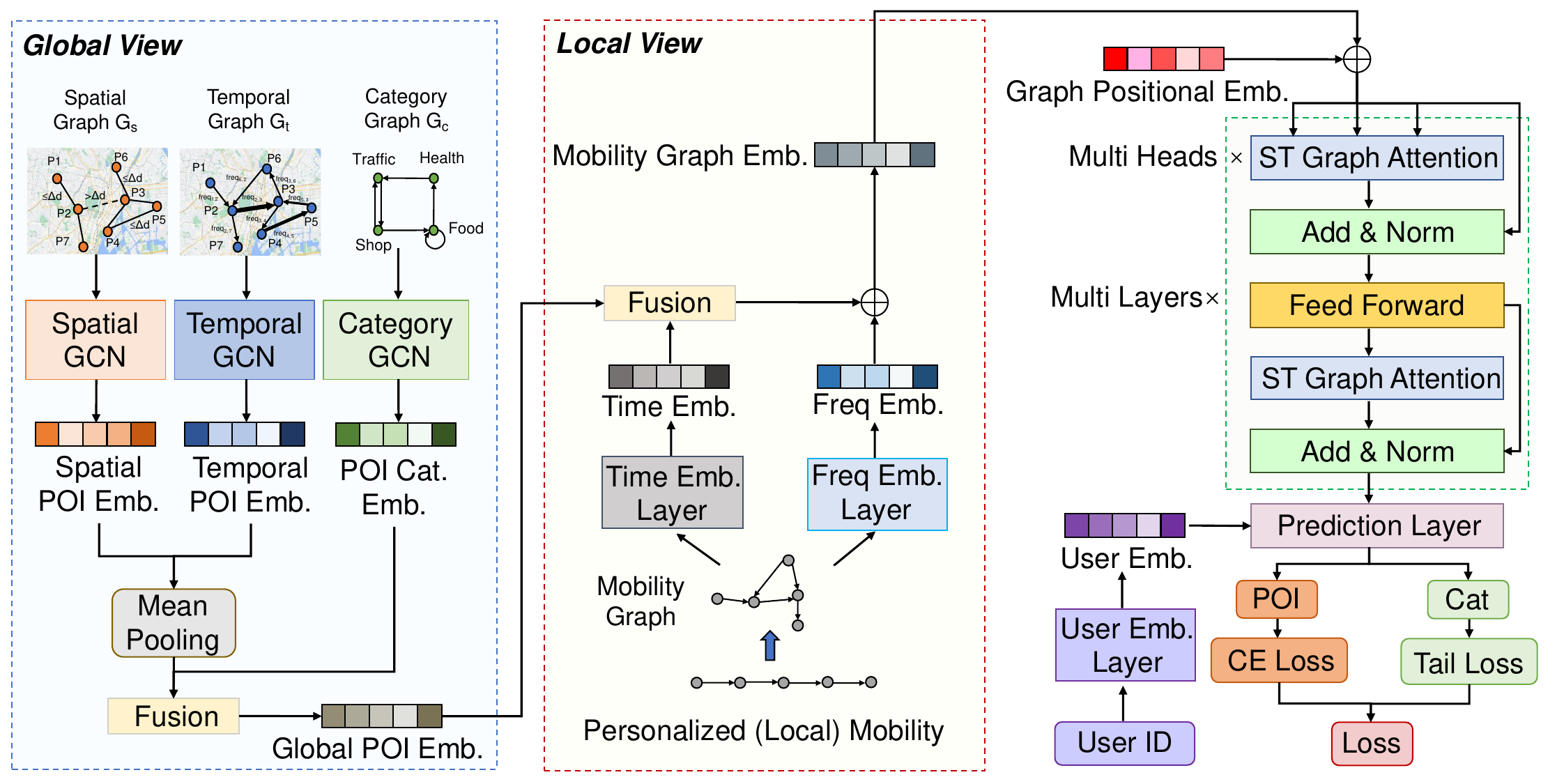}
\caption{The overview of our proposed \textbf{\underline{Mob}}ility \textbf{\underline{G}}raph \textbf{\underline{T}}ransformer (MobGT).}
\label{fig:STGTrans}
% \vspace{-0.5cm}
\end{figure*}

% \subsection{Model Overview}
% \label{sec:4_model}
In Fig.~\ref{fig:STGTrans}, we present the overall architecture of \textbf{\underline{Mob}}ility \textbf{\underline{G}}raph \textbf{\underline{T}}ransformer (MobGT), which integrates diverse contextual perspectives.
Initially, we define three global graphs from distinct viewpoints, capturing the geographic and transition information between POIs and POI categories to model the mobility patterns of all users. 
% This comprises two components:
% \begin{enumerate}
% \item We use GNN to train POI embeddings from the spatial and temporal dimensions, incorporating inherent geographic, category, and access frequency features of POIs.
% \item The changes in POI categories can also provide significant insights, as each POI is accompanied by a unique POI category, thus we propose a transition graph between different POI categories and train the POI category embedding by GNN.
% \end{enumerate}
Subsequently, we leverage more precise mobility contextual information to learn user, time, and POI frequency embeddings.
% \begin{enumerate}
% \item User embeddings personalize recommendations for users in the local view.
% \item Time embeddings capture users' temporal preferences by learning their movement patterns during specific time periods.
% \item Category embeddings complement POI embeddings from a macro perspective, as they have fewer categories and more stable transition patterns.
% \item POI frequency embeddings help mitigate the long-tail problem, as some POIs are visited less frequently.
% By utilizing POI frequency embeddings and GradientLoss, \projtitle can effectively alleviate the long-tail problem.
% \end{enumerate}
Finally, we input POI embeddings generated from the global and the local graph, as well as user, category embeddings, and other contextual information into our proposed Spatial-Temporal (ST) graph attention model with individual mobility graphs to learn the mobility characteristics of each user.
% For prediction, we adopt a multi-task architecture to provide accurate POI recommendations and mitigate the long-tail problem.

% \vspace{-0.3cm}

\begin{figure}[!tpb]
\centering	
\includegraphics[width=0.48\textwidth]{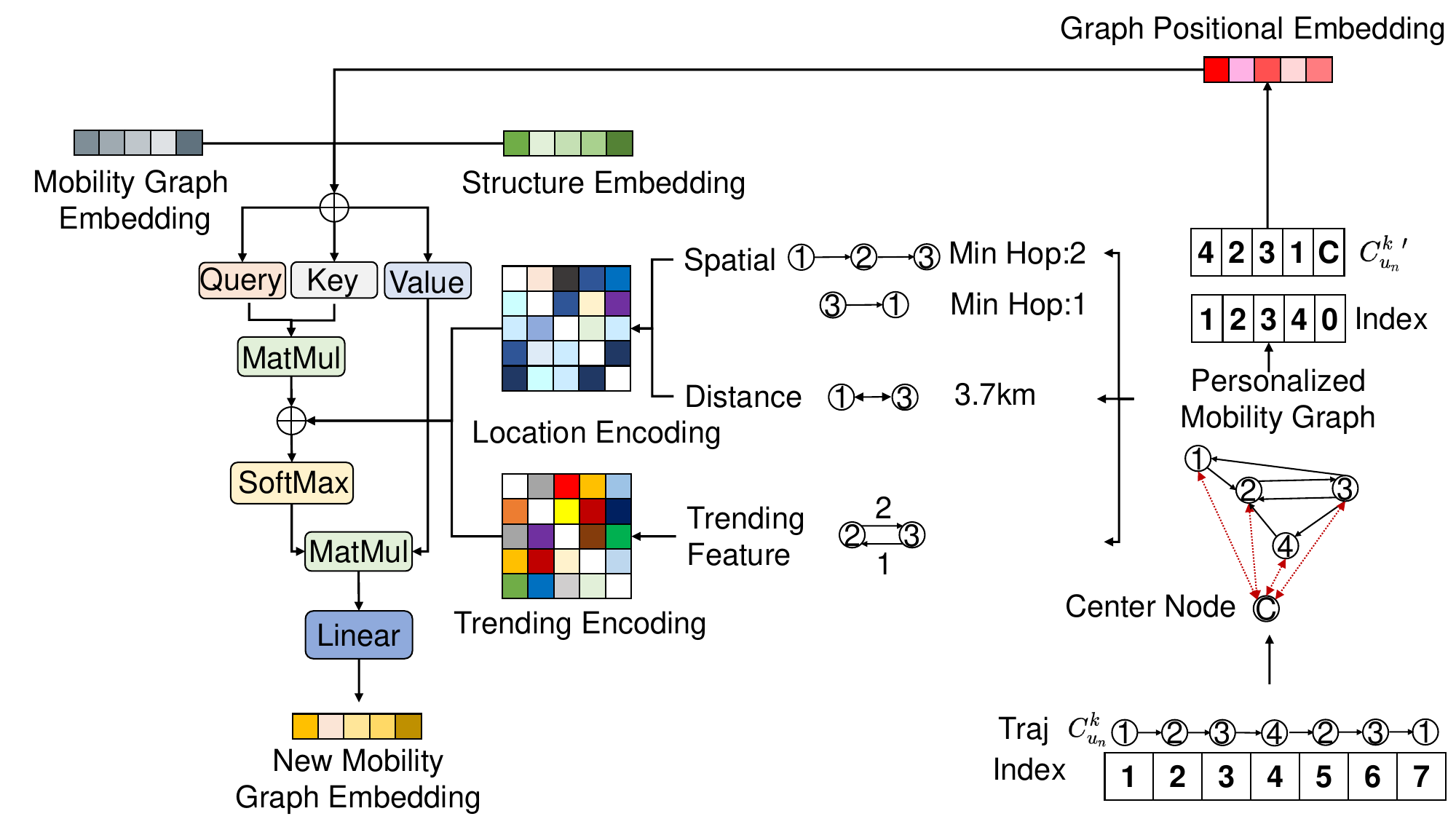}
\caption{The overview of the ST Graph Attention module.}
\label{fig:STGT}
% \vspace{-0.5cm}
\end{figure}

\subsection{POI Embedding via Global Graph}
\label{sec:4_global}
% \textbf{Global Graph.} 
% Popular POI recommendation models often focus only on personalized recommendations based on individual users' mobility patterns, without exploring similarities between different users' trajectories. 
% In addition, the inter-category transition relationships are crucial for predicting the next POI. 
% For example, after check-in a restaurant, a user is more likely to visit an entertainment or shopping mall, rather than go to another restaurant. 
% To address this, we propose three types of global graphs, from spatial, temporal and categorical perspectives, to learn POI embeddings. 
In this section, we propose three types of global graphs, from spatial, temporal, and categorical perspectives, to learn POI embeddings. 

\textbf{Global Spatial Graph.} We build an undirected POI-POI graph where the set of nodes (POIs) is represented by $P$ and each undirected edge between POIs is represented by $E_{sg}$. 
If two POIs are adjacent geographically, that is, the Haversine distance $\Delta x$ between the two POIs is below a threshold, i.e., 2.5 km, we add an edge between them in $E_{sg}$.
It should be noted that since the Earth is a three-dimensional sphere rather than a two-dimensional plane, we use the Haversine distance rather than the Euclidean distance to calculate the distance between two POIs.

\textbf{Global Temporal Graph.} Then we propose a directed POI-POI graph with nodes set $P$ and directed edges $E_{tg}$, where edge weight indicates the frequency of a POI pair in all trajectories.
% Then we propose a directed POI-POI graph where each directed edge between POIs is represented by $E_{tg}$. The set of nodes is represented by $P$, and the edge weight signifies the number of times a POI pair appears in all trajectories.

\textbf{Global Category Graph.} Finally, we propose a global POI category graph which is a directed Cat.-Cat. graph, represented by $G_{cg}=(V_{cg}, E_{cg})$, where $V_{cg}$ is the set of categories that each POI belongs to. 
We connect each directed category pair based on all users' trajectories. 
The edge weight between each pair is the frequency of category pair occurrences.

To obtain POI embeddings, we employ GCN to learn the relevant contextual features from the three global graphs.
For the spatial $e_{s}$ and temporal POI embeddings $e_{t}$, we first feed them into a pooling layer to obtain the aggregated POI embedding $e_{st}$. 
We then fuse the POI and category embeddings $e_{c}$ using the following approach:
\begin{equation}
e_{p}=\sigma(W_{stc}[e_{st}||e_{c}]+b_{stc}),
\label{eq:GCN3}
\end{equation}
where $||$ denotes the concatenation and $\sigma$ is the Leaky ReLu activation function.

\subsection{Mobility Embedding via Local Graph}
\label{sec:4_local}
The global graph of learning POI and category embeddings aggregates information from all users and lacks the learning of users' preferences. 
% Next, we propose our local view module to enrich the semantic information of POIs by fusing POI embeddings, time embeddings, and POI frequency embeddings.
Next, we propose our local graph to enrich the semantic information of POIs by fusing different contextual embeddings.

For the POI recommendation task, user check-in behavior has a strong dependence on time~\cite{yuan2013time,he2019inferring,wang2021forecasting}. 
As shown in Fig.~\ref{fig:data_feature_2} for the Gowalla and Toyota datasets, users are more active during rush hours, so we need a time-aware model that can better explore users' mobility patterns. 
For example, a white-collar worker's mobility on workdays is almost always from 7-9 a.m. to the station and from 5-7 p.m. back from the station. 
In Fig.~\ref{fig:data_feature_3}, in the Foursquare-TKY dataset, for two different POI categories "Food \& Drink" and "University", "University" is visited more frequently during the day than "Food \& Drink", so we need to consider how to make the model learn to capture the relationship between POI and time patterns, i.e., recommend "University" as users' destination rather than "Food \& Drink" at 9:00 AM. 
% because most users are unlikely to visit "Food \& Drink" in the morning.
% Also, it is evident that direct input of continuous check-in time into the model presents certain challenges. 
% To address this, we discretized the check-in time using 48 bins, i.e. dividing each day into 48 temporal intervals.
% To build the connection between POIs and check-in time, we use an embedding layer $f_{time}$ to generate low-dimensional vector representations $e_{ti} \in \mathbb{R}^{d_{ti}}$ for check-in time $ti$. 
To capture the relationship between POIs and time patterns, we discretize check-in times into 48 intervals per day and utilize an embedding layer $f_{time}$ to generate low-dimensional vector representations $e_{ti} \in \mathbb{R}^{d_{ti}}$ for each time interval $ti$.
Specifically, the time embedding $e_{ti} \in \mathbb{R}^{d_{ti}}$ can be denoted as:
\begin{equation}
e_{ti}=f_{time}(ti).
\label{eq:localview1}
\end{equation}
Next, we utilize the same approach as the fusion of POI and POI category to aggregate the information for POI embedding and time embedding and get an aggregation POI embedding $e_{pti} \in \mathbb{R}^{d_{p}+d_{c}+d_{ti}}$.

% \begin{equation}
% e_{pti}=\sigma(W_{pti}[e_{p}||e_{ti}]+b_{pti}),
% \label{eq:localview2}
% \end{equation}
% where $e_{pti} \in \mathbb{R}^{D}, D=d_{p}+d_{c}+d_{ti}$. 
From Fig.~\ref{fig:data_feature_2}, we observe that the frequency of POI visits varies significantly across different time periods, 
Also, several studies~\cite{yang2022getnext,lim2020stp,li2019context} have highlighted the importance of frequency for POI recommendation, so it is important to consider using this frequency information to enrich the semantic representation of POIs.  
Therefore, we use an embedding layer $f_{freq}$ to train a POI frequency $freq$ embedding $e_{f} \in \mathbb{R}^{d_{p}+d_{c}+d_{ti}}$, which can be represented as:
\begin{equation}
e_{f}=f_{freq}(freq).
\label{eq:localview3}
\end{equation}
Finally, we fuse the POI embedding $e_{pti}$ and POI frequency embedding $e_{f}$ by element-wise addition operation:
\begin{equation}
e_{o}=e_{pti}+e_{f}.
\label{eq:localview4}
\end{equation}

To predict the next POI to be visited, we convert each check-in mobility into a graph and feed it into our proposed spatial-temporal (ST) graph attention module. 
Inspired by~\cite{ishiguro2019graph}, we add a center node $v_{center}$ to aggregate the information of the entire graph. 
The center node is connected to all nodes in the graph in an undirected manner, enabling multi-hop propagation of information. 
This allows the center node to capture the information of all nodes and serve as a relay node. 
As the sequence model is propagated hop-by-hop when passing information, the center node can facilitate multi-hop propagation of information among all nodes.

% After obtaining the POI embeddings, we need to convert each check-in trajectory $C=\{k_{1}, k_{2}, \cdots, k_{n}\}$, into a graph and feed it into the proposed spatial-temporal(ST) graph attention module.
% As our ultimate goal is to predict the next possible POI to be visited based on each trajectory subgraph, we were inspired by the work of global nodes in~\cite{ishiguro2019graph,ying2021transformers} and added a center node $v_{c}$ when converting the trajectory sequence into a graph $G_{C_{u}^{tr}}$ to aggregate the information of the entire graph. 
% The center node $v_{c}$ is connected to all nodes $v_{p} \in V_{C_{u}^{tr}}$ in the graph $G_{C_{u}^{tr}}$ in an undirected manner, meaning that all nodes in the graph can be accessed from the center node.
% The center node can not only capture the information of all nodes but can also be used as a relay node to enable all nodes to achieve multi-hop propagation of information, as the sequence model is propagated hop-by-hop when passing information. 
% Furthermore, when predicting, the center node embedding that integrates all information in the trajectory graph is the POI embedding that we want to predict, as it combines all information in the entire trajectory graph.

% \vspace{-0.3cm}
\subsection{Spatial-Temporal Graph Attention}
\label{sec:4_ST}
In this section, we introduce our proposed Spatial-Temporal (ST) graph attention module illustrated in Fig.~\ref{fig:STGT} for predicting the user's next POI based on the embedded mobility graph. 
% In this section, we first introduce our proposed ST graph attention model for predicting the next POI based on the global and local views learned from the user trajectory graph. 
% It is a graph-based model that learns the topology, location, and transition information between nodes, rather than relying on sequential information. 
% Secondly, we elaborate on the multi-task decoder component. 
% Finally, we discuss how to alleviate the long-tail problem by combining multiple loss functions.

% \subsubsection{Structure Encoding}
% \label{sec:4_structure}
\subsubsection{Structure Encoding} After converting mobility sequences into graphs, we can utilize the degree of nodes to enhance the model's capacity to capture the structure information of the entire graph.
For example, when a POI is frequently visited in a subgraph of mobility, it is likely to be the user's workplace or home address. 
However, if the model only depends on attention mechanisms to calculate their potential connections is far less effective than directly inputting graph structure information into the model.
% Therefore, we propose to define three new structural encodings to enrich the semantic information of POIs in the trajectory graph through explicit graph structure input:
Inspired by~\cite{ying2021transformers}, we propose to define three new structural encodings to enrich the semantic information of POIs in the mobility graph through explicit graph structure input:
\begin{equation}
h_{s}=e_{o}+e_{deg^{-}(v_{s})}+e_{deg^{+}(v_{s})}+e_{pos(v_{s})},
\label{eq:stgraph1}
\end{equation}
where $h_{s}, e_{deg^{-}(v_{s})}, e_{deg^{+}(v_{s})}, e_{pos(v_{s})}\in \mathbb{R}^{d}$ are all learnable embeddings.
% where $h_{s}\in \mathbb{R}^{d}, e_{deg^{-}(v_{s})}\in \mathbb{R}^{d}, e_{deg^{+}(v_{s})}\in \mathbb{R}^{d},$ and $e_{pos(v_{s})}\in \mathbb{R}^{d}$ are all learnable embeddings. 
Here, $e_{deg^{-}(v_{s})}$ and $e_{deg^{+}(v_{s})}$ refer to the embeddings of node in-degree and out-degree, respectively, and $e_{pos(v_{s})}$ is the embedding of the node's position.
% Taking the Transformer~\cite{vaswani2017attention} as an example, sequence models added a position embedding component to describe the order of the sequence input. 
% However, when we convert a sequence to a graph if cycles occur due to the repetition of nodes, it unavoidably leads to confusion in position information. 
% Therefore, due to the nature of the graph structure itself, we can only encode the position of each POI's last appearance in the trajectory.
% Specifically, for the input sequence $\{1, 2, 3, 4, 2, 3, 1\}$, we can only retain $\{4, 2, 3, 1\}$ as the input to the graph, and the specific position input is shown in Figure~\ref{fig:STGT}. 
Meanwhile, when we convert a sequence to a graph if cycles occur due to the repetition of nodes, so we encode the position of each POI's last appearance in the trajectory.
For introducing the index position of nodes $pos(v_{s})$, we use a trainable position matrix $W_{pos}$ to generate a positional embedding $e_{pos(v_{s})}$ for each input node.

% \begin{equation}
% e_{pos(v_{s})}=W_{pos}pos(v_{s}),
% \label{eq:stgraph2}
% \end{equation}
% where $W_{pos}$ is a trainable matrix, and $pos(\cdot)$ represents the index of node $v_{s}$.

% \subsubsection{Location Encoding}
% \label{sec:4_location}
\subsubsection{Location Encoding} The difference between POI recommendation and traditional recommendation tasks lies in its unique geographical information, which we have considered from a global perspective. 
% Generally, people tend to visit POIs that are closer to their current location. 
% However, each person's mobility habits are different, and our dataset contains two modes of movement: human movement and vehicle movement. 
% When people move by car, distance is no longer a problem, and users will choose places that better match their interests. 
Therefore, we also encode two types of location information for nodes in the mobility subgraph. 
Firstly, we calculate the shortest hop (sh, for human behavior) count between each POI. 
Subsequently, we use the Haversine function~\cite{winarno2017location} to compute the true distance
% (dist, for distinguishing car and human behavior) 
between each POI. 
For traditional GNNs, the model typically focuses on its first-order neighbors, such as GAT~\cite{velivckovic2017graph}. 
However, by calculating the shortest hop count, we introduce high-order neighbor relationships between nodes into the model. 
Specifically, for any two nodes $v_{s}$ and $v_{t}$, we define:
\begin{equation}
Local\_Attn(v_{s}, v_{t})=\frac{(v_{s}Q)(v_{t}K)^T}{\sqrt{d}}+l_{sh(v_{s}, v_{t})}+l_{dist(v_{s}, v_{t})},
\label{eq:stgraph3}
\end{equation}
where $l_{sh(v_{s}, v_{t})}$ is a learnable scalar indexed by $sh(v_{s}, v_{t})$, and $l_{dist(v_{s}, v_{t})}$ is a learnable scalar trained by binning algorithm. 
Furthermore, Q and K are the corresponding trainable attention weight matrices.
As the geographical distance between two nodes is a continuous floating-point number, it would be challenging to generate trainable embeddings directly, and the distance between two POIs is also randomly distributed. 
To make the number of POIs in each bin more evenly distributed when discretizing, we use the Freedman-Diaconis formula~\cite{chuan2019comparison} to calculate the number of bins, we denote the number of bins $b$ as follows:
\begin{equation}
b=\frac{max(dist)-min(dist)}{2 \times \frac{IQR}{\sqrt[3]{n}}},
\label{eq:stgraph4}
\end{equation}
where IQR is the interquartile range, and $n$ is the number of samples.

% \subsubsection{Trending Encoding}
% \label{sec:4_trending}
\subsubsection{Trending Encoding} For a given sequence $\{1, 2, 3, 4, 2, 3, 1\}$, we transform it into a graph structure and obtain node set $\{4, 2, 3, 1\}$. 
% So far, we have not considered the transition tendencies between nodes. 
For POI pairs such as $(2, 3)$ that occur multiple times, we should incorporate these high-frequency POI pairs as input information in our encoding component. 
% Therefore, we suggest treating visit trends as edge features to better describe the type characteristics between nodes. 
% For example, the node pair $(2, 3)$ may correspond to the trajectory from the subway station to an office. 
% Furthermore, as our model has already introduced higher-order features in the Location Encoding part, prior work has not effectively captured feature information beyond single-hop nodes. 
% To address this issue, we propose a graph-based Trending Encoding method. 
Furthermore, previous studies did not effectively capture feature information beyond single-hop nodes. 
To address this issue, we propose a graph-based Trending Encoding method.
Specifically, for any node pair $(v_{s}, v_{t})$, similar to Location Encoding, we can find the shortest hop $sh_{(v_{s},v_{t})}=(d_{1}, d_{2}, \cdots, d_{N})$ to aggregate all higher-order relationships. 
Along this shortest hop, we aggregate all edge features as our Trending Encoding, which can be expressed as follows:
\begin{equation}
T_{st}=\frac{1}{N}\sum_{n=1}^{N}W_{d}d_{n},
\end{equation}
where $W_{d}$ is a trainable matrix. 
Additionally, to introduce Trending Encoding into the two encoding methods mentioned above without disrupting the original data flow, we treat it as the bias term in the attention module. 
Thus, our ST Graph Attention can be denoted as:
\begin{equation}
ST\_Attn(v_{s}, v_{t})=\frac{(v_{s}Q)(v_{t}K)^T}{\sqrt{d}}+l_{sh(v_{s}, v_{t})}+l_{dist(v_{s}, v_{t})}+T_{st}.
\end{equation}

Finally, we use the encoder structure of the Transformer to input the node embeddings that have passed through the ST Graph Attention module into several feed-forward layers (FFL) for normalization. 
% The final trainable layer $l$ can be expressed as:

% \begin{equation}
% l^{(n)}=FFL(Norm(ST\_Attn(\lambda(l^{(n-1)}))+l^{(n-1)}))+ST\_Attn(\lambda(l^{(n-1)}))+l^{(n-1)},
% \end{equation}
% where $\lambda$ means layer normalization.

% \subsubsection{Decoder and Prediction}
% \label{sec:4_prediction}
% \vspace{-0.3cm}
\subsection{Prediction and Optimization}
In this section, we describe how to utilize user information, which can improve the effect of personalized recommendations, and introduce the design of the prediction layer and optimization methods.
% \textbf{Decoder and Predictor.}

\subsubsection{User Embedding and Prediction Layer} The training layer learns the user's transition patterns from the first-order and higher-order features embedded in the input mobility subgraphs. 
However, to further incorporate user information $v_{s}(user)$ into the model explicitly and provide more personalized POI recommendation services, we introduce user embedding $e_{u}$ by a trainable matrix $W_{u}$.
% \begin{equation}
% e_{u}=W_{u}(v_{s}(user)),
% \end{equation}
% where $W_{u}$ is a trainable matrix. 
We also denote the mobility embedding output by the training layer as $e_{tl}$. 
We use the same fusion method in a global graph to concatenate the user embedding and the mobility embedding to form a representation $e_{utl}$.
% \begin{equation}
% e_{utl}=\sigma(W_{utl}[e_{u}||e_{tl}]+b_{utl}),
% \end{equation}
% where $W_{utl}$ is a trainable matrix, and $b_{utl}$ is the bias term. 
% To achieve our ultimate goal of predicting the user's next visited location, 
% we adopt a multitask approach using 
To predict the user's next visited location, we adopt two different multilayer perceptron models (MLP) as prediction layers to predict the next POI and POI category, respectively.
\begin{equation}
\begin{aligned}
&\hat{y}_{poi}=W_{poi}e_{ctl}[v_{center}]+b_{poi}, \\
&\hat{y}_{cat}=W_{cat}e_{ctl}[v_{center}]+b_{cat},
\end{aligned}
\end{equation}
where $W_{poi}$ and $W_{cat}$ are the weighted matrices of the MLP, and $b_{poi}$ and $b_{cat}$ are their corresponding bias terms.
$e_{ctl}[v_{center}]$ refers to using the center node we mentioned as the training target, which contains all semantic information (POI, POI category, visited frequency, etc.) of the user's next possible visited POI. 
% After decoding steps, we can obtain the information we want to predict, such as the next POI or POI category.

% \subsubsection{Loss}
% \label{sec:4_loss}
\subsubsection{Optimization} 
We employ cross-entropy as the main loss function $\mathcal{L}_{ce}$ for POI prediction. 
However, as the number of POI categories is significantly smaller than the number of POIs (i.e., the long-tail problem), the model may face challenges in learning the features of certain POIs with limited data, resulting in model performance degradation.
% To ensure the fairness of our experiments, we preserved the original data and refrained from adopting any special preprocessing methods. 
To alleviate this, inspired by~\cite{chen2022gradtail}, 
we introduce an extra task of predicting the POI category and propose a Tail Loss $\mathcal{L}_{tail}$ as follows:
\begin{equation}
\begin{split}
\mathcal{L}_{tail} = \frac{1}{n} \sum_{i=1}^{n} \bigg[ &-\alpha(1-\sigma(x_{i}))^k y_{i} \log(\sigma(x_{i})) \\
&- \beta\sigma(x_{i})^k(1 - y_{i})\log(1-\sigma(x_{i}))\bigg],
\end{split}
\end{equation}
where $x_{i}$ is the model's predicted output for sample $i$, $y_{i}$ is the true class label for sample $i$, $n$ is the number of samples, $\sigma$ means the sigmoid activation function, $\alpha$ and $\beta$ are hyperparameters used to control the weight of the positive and negative losses, respectively, and $k$ is an exponent used to control the sensitivity to the tail of the probability distribution. 
% Furthermore, to increase the impact of POI categories in training, we multiplied their loss values by a hyperparameter $\lambda$. 
% the scale of their losses is relatively small. 
% To increase the impact of POI categories in training, we multiplied their loss values by 10. 
The final loss can be denoted as:
\begin{equation}
\mathcal{L}=\mathcal{L}_{ce}+\lambda \mathcal{L}_{tail},
\end{equation}
where $\lambda$ is a balancing factor for the two losses. In our study, $\alpha$, $\beta$, $k$, and $\lambda$ are set to 0.2, 1, 1.2, and 10.

\section{Experiments}
\subsection{Datasets}\label{sec:5_Datasets}

\label{sec:5_experiments}
% \small
% \vspace{-0.5cm}
\begin{table}[h]
\centering
\caption{Basic Statistics of Datasets}
\label{tab:data-statstic}
\resizebox{0.48\textwidth}{!}{
\begin{tabular}{lccc}
% \begin{tabular}{lcccc}
\toprule
Dataset    & Gowalla-NV & Foursquare-TKY & Toyota-TKY
%& \multicolumn{1}{l}{Foursquare-NYC}%
\\ \midrule
% Duration  & Apr. 2012- Feb. 2013  & Feb. 2009-Oct. 2010  & May 2021-Apr. 2022 \\
% \#Users  & 2,261 & 1,080  & 995  \\
% \#POIs   & 7,855 & 3,679  & 8,011  \\
% \#Categories  & 291 & 253 & 49  \\
% \#Check-Ins   & 363,163 & 87,828 & 539,755  \\
% \#Trajectories & 44,669 & 6,869 & 51,880  \\
% \#POIs / User & 38 & 35 & 106 \\ 

Duration  &2009.02$\sim$2010.10  & 2012.04$\sim$2013.02  & 2021.05$\sim$2022.04 \\
\#Users  &1,080  & 2,261  & 995  \\
\#POIs   &3,679  & 7,855 & 8,011  \\
\#Categories  &253  & 291 & 49  \\
\#Check-Ins   &87,828  & 363,163 & 539,755  \\
\#Trajectories &6,869  & 44,669 & 51,880  \\
\#POIs / User &35  & 38 & 106 \\ 

\bottomrule 
\end{tabular}
}
% \vspace{-0.5cm}
\end{table}

Three datasets were utilized in our experiments to assess the performance of our approach. 
% These datasets were collected from human mobile check-in data and vehicle GPS data.
The first two datasets were collected from two public LBSN (Location-Based Social Network) platforms: Gowalla-Nevada~\cite{liu2014exploiting} and Foursquare-TKY~\cite{yang2014modeling}. 
% The Foursquare-TKY dataset contains check-ins made by residents of Tokyo on the Foursquare app from April 2012 to February 2013, while the Gowalla-NV dataset records check-ins made by users in the Nevada region on the Gowalla platform from February 2009 to October 2010. 
Each record in these datasets includes the user ID, check-in POI, timestamp, latitude, longitude, and POI category.
The third dataset was collected from the GPS systems of Toyota cars. 
% The Toyota dataset was collected from the GPS systems of Toyota cars from May. 2021 to Apr. 2022. 
The dataset records the vehicle data of the Tokyo metropolitan area mainly covering the Harumi, Kichijyoji, Kitasenjyu, Musasukosugi, and Tamapuraza areas. 
We selected the top 200 most active users from these five regions to form the entire dataset. 
In addition, the dataset does not include taxi users.
Each record in the dataset contains the user ID, timestamp, latitude, and longitude information, with POI and POI category information obtained through data mapping based on the Toyota database.
% The system recorded the latitude and longitude information, as well as the timestamp, of the current location whenever the car was started or the engine was turned off. 
% It is worth mentioning that the GPS coordinates have different mappings with POIs, which were obtained from the Toyota TMI database. 
% The dataset records the vehicle data of the Tokyo metropolitan area from May 2021 to April 2022.
% mainly covering the Harumi, Kichijyoji, Kitasenjyu, Musasukosugi, and Tamapuraza areas. 
% Each record in these datasets includes the user ID, check-in POI, timestamp, latitude, longitude, and POI category.
% Each record in the dataset contains the user ID, timestamp, latitude, and longitude information, with POI information and type obtained through later data mapping.
We list the main statistics in Table~\ref{tab:data-statstic}.

\begin{figure}[h]
	\centering	
	\subfigure[The cumulative distribution function of location visit frequency of all mobility records.]{
		\includegraphics[width=0.22\textwidth]{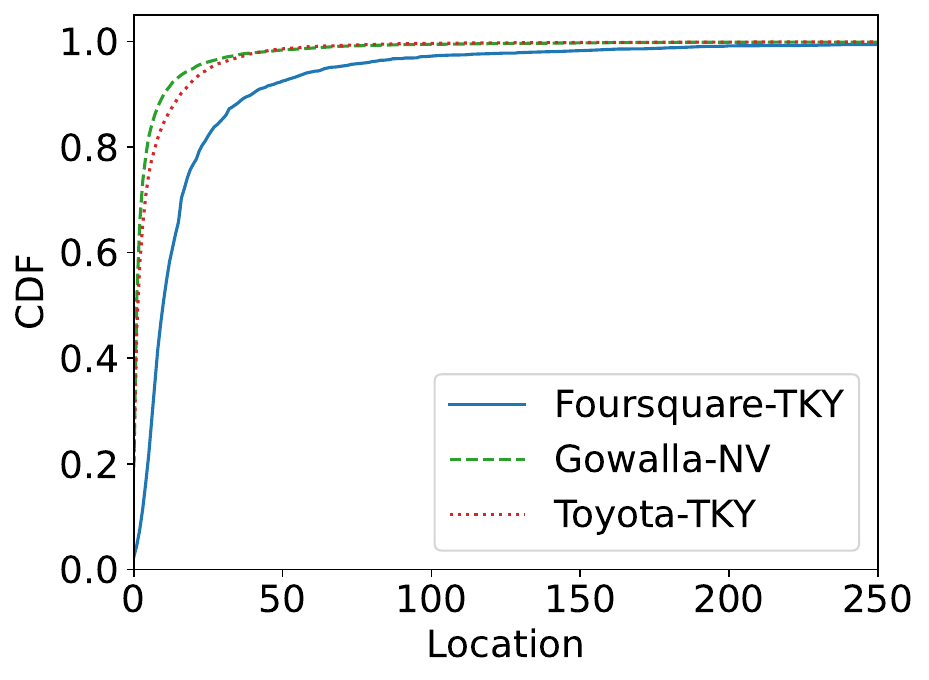}
		\label{fig:data_feature_1}
	}\ \ \
	\subfigure[The probability distribution function of time slots of all mobility records.]{
		\includegraphics[width=0.22\textwidth]{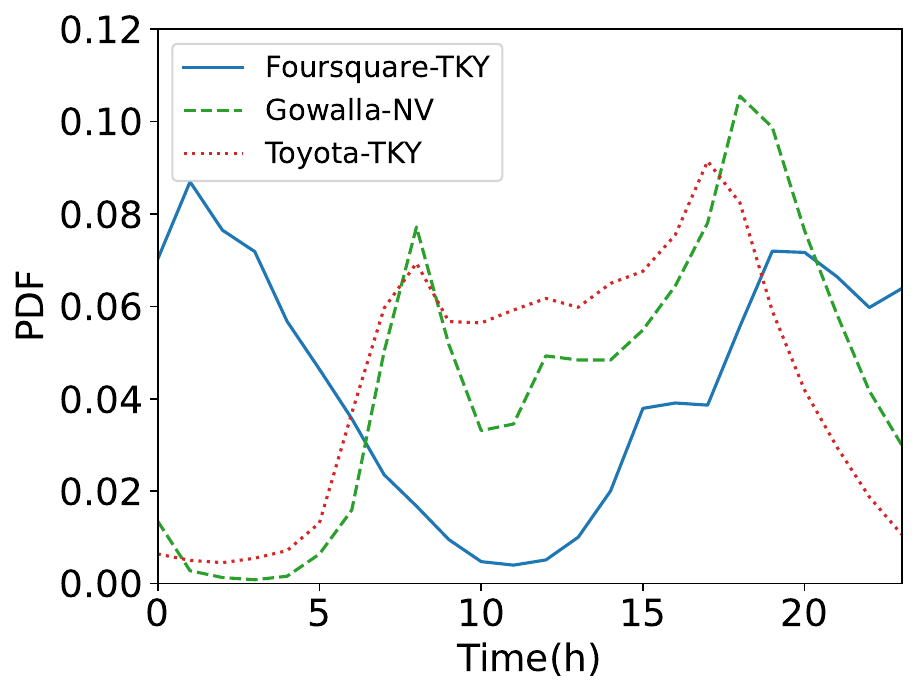}
		\label{fig:data_feature_2}
	}
 	\subfigure[The check-in frequency of ``Food \& Drink'' and ``University'' POI categories at the hourly level.]{
		\includegraphics[width=0.24\textwidth]{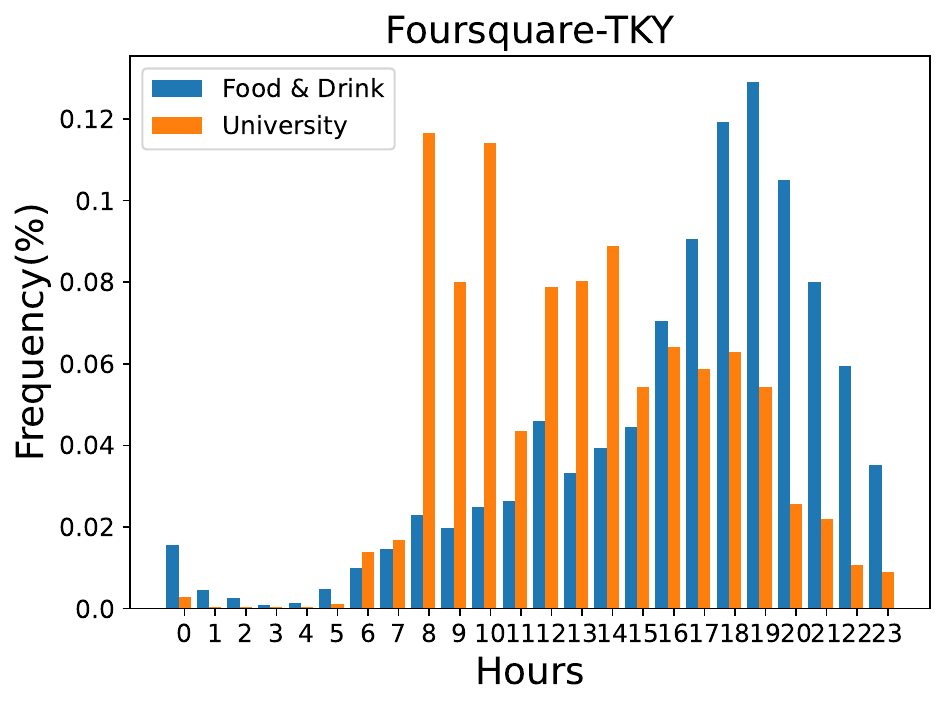}
		\label{fig:data_feature_3}
	}
	\caption{Spatial-temporal features detailed information of three mobility datasets.}\label{fig:data_feature}
 \vspace{-0.5cm}
\end{figure}
\begin{figure}[!tpb]
	\centering	
	\subfigure[Gowalla-NV]{
		\includegraphics[width=0.22\textwidth]{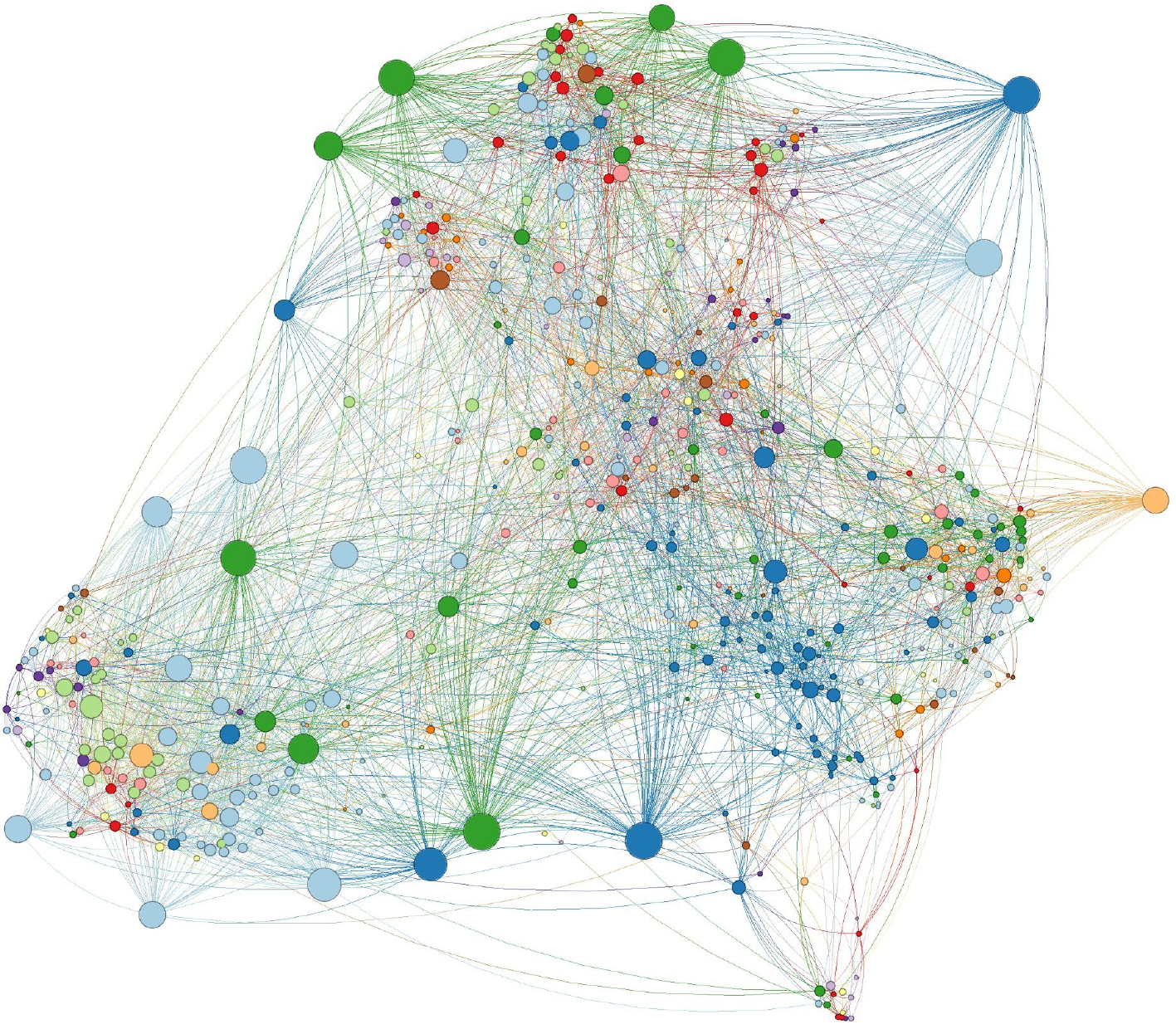}
		\label{fig:data_visual_1}
	}\ \ \
	\subfigure[Foursquare-TKY]{
		\includegraphics[width=0.22\textwidth]{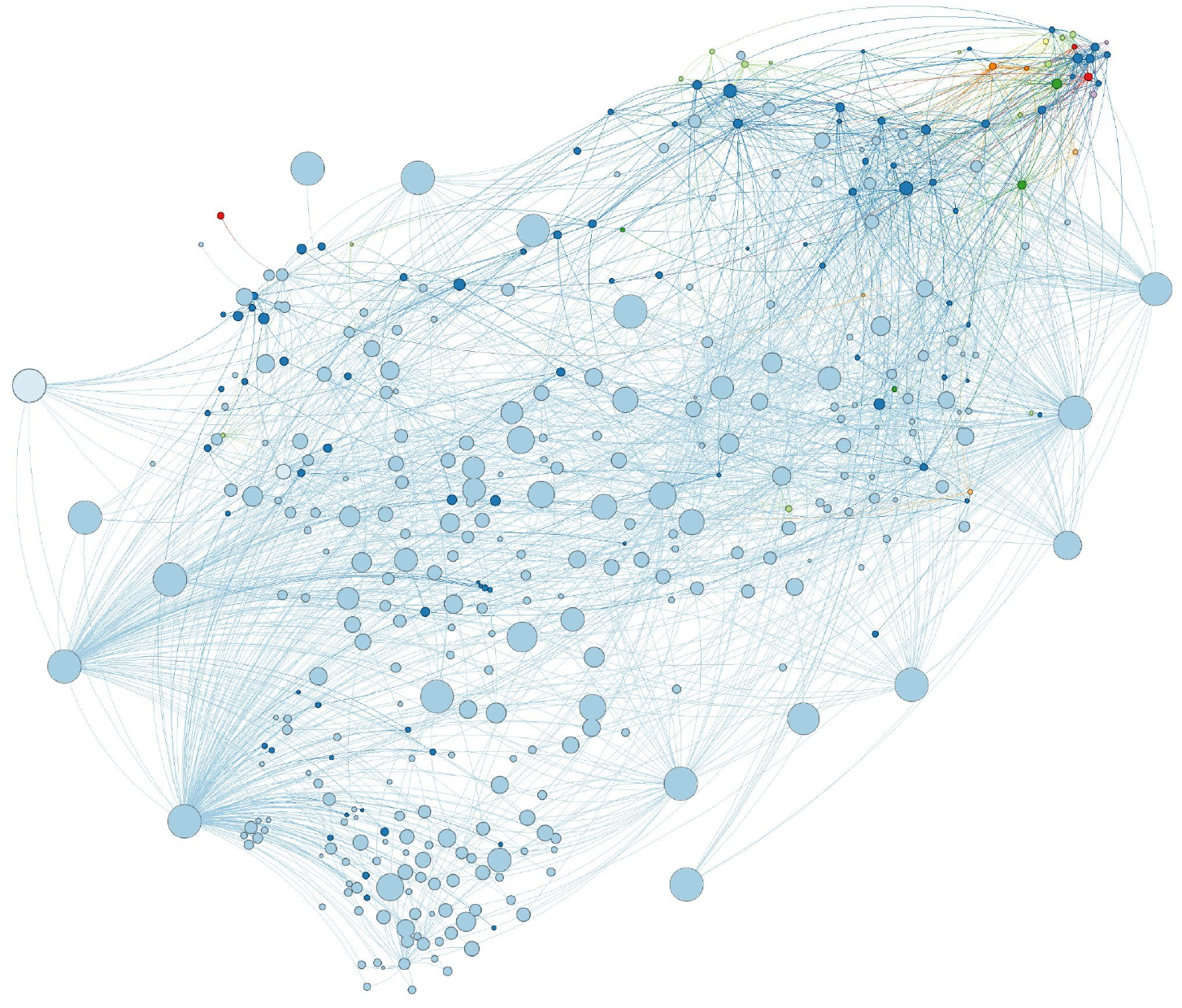}
		\label{fig:data_visual_2}
        }
        \subfigure[Toyota-TKY]{
		\includegraphics[width=0.22\textwidth]{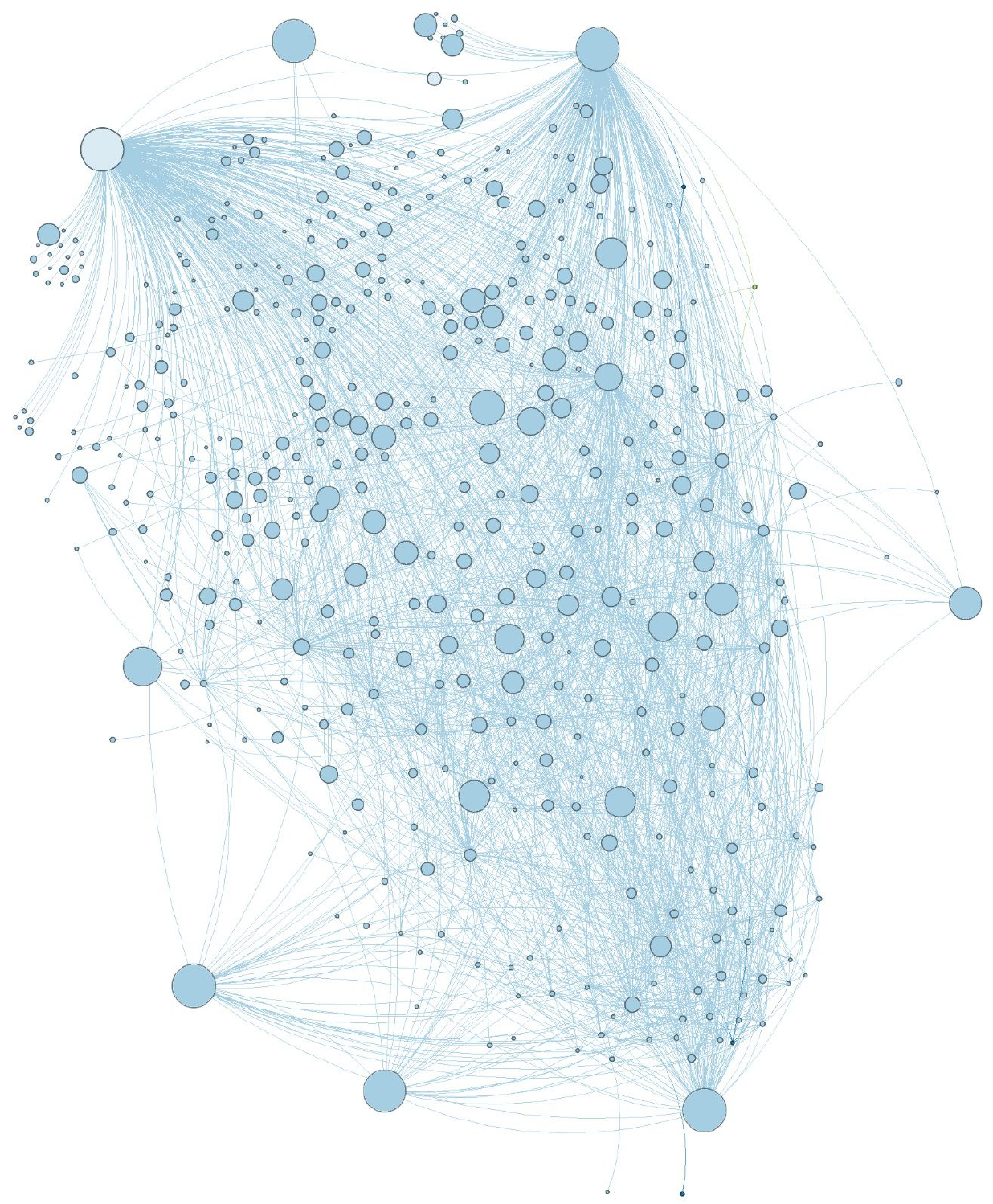}
		\label{fig:data_visual_3}
	}
	\caption{Partial mobility graph of all users on the three datasets. We use undirected graphs for better visualization. A larger node size represents a larger degree.}\label{fig:data_visual}
 % \vspace{-0.5cm}
\end{figure}

% For all three datasets, we eliminate POIs with fewer than 10 check-ins and remove users with fewer than 2 check-ins. 
We partition the check-in records into multiple trajectories based on a 1-day time interval, ensuring that each trajectory contains a minimum of 3 check-in POIs. 
Subsequently, we split the data into training and test sets based on the chronological order of check-ins, with the first 80\% of the trajectories comprising the training set, while the remaining 20\% data make up the test set.
The spatial-temporal characteristics of the datasets are depicted in Fig.~\ref{fig:data_feature}. 
Specifically, Fig.~\ref{fig:data_feature_1} illustrates that the majority of POIs are visited fewer than 50 times. 
Fig.~\ref{fig:data_feature_2} depicts that the users of the Foursquare dataset exhibit heightened activity during nighttime, whereas the users of Gowalla and Toyota datasets exhibit peak activity during the morning hours between 7-9 and evening hours between 17-19, which aligns with typical human activity patterns.
% For Fig.~\ref{fig:data_feature_3}, we have already introduced it in Sec.~\ref{sec:4_local}, which reflects the popularity of POI category being visited in different time periods.
Fig.~\ref{fig:data_visual} depicts the distribution of the three datasets after visualization. 
We can observe that in the Toyota dataset, there is a noticeably higher number of hot visited points compared to the other two datasets. 
Here, "hot visited points" refer to the frequency of node visits, with the size of the nodes in the graph representing the visitation frequency. 
On the other hand, this also reflects that the movement patterns of vehicles are more regular than those of humans, which is one of the reasons our model performs better on the Toyota dataset.

% \subsection{Baselines}
% \label{sec:5_Baselines}
% We consider conducting the following traditional and state-of-the-art models as the baselines.
% \begin{itemize}
% \item MC~\cite{gambs2012next}is a conventional recommendation model that forecasts the subsequent item based on the probabilities derived from the Markov Chains.

% \item LSTM~\cite{hochreiter1997long} is a variation of the RNN model designed to effectively model sequence through the simultaneous consideration of both long-term and short-term dependencies.

% \item ST-RNN~\cite{liu2016predicting} extends the traditional RNN by incorporating the transformation matrix of temporal and spatial features.

% \item DeepMove~\cite{feng2018deepmove} captures sequential patterns by integrating multiple attention mechanisms to model both the long-term and short-term interests of users.

% \item Flashback~\cite{yang2020location} is an RNN-based model that places a strong emphasis on modeling historical check-in data.

% \item LSTPM~\cite{sun2020go} utilizes a geo-nonlocal LSTM architecture to model the evolving patterns of long-term interests and a geo-dilated LSTM to capture the geographic dependencies of short-term interests. 

% \item STAN~\cite{luo2021stan} employs an attention mechanism to learn the spatial-temporal relationships between non-adjacent POIs in the check-in sequence.

% \item GETNext~\cite{yang2022getnext} is a Transformer-based model that utilizes all user mobility patterns to represent POIs.
% GETNext is the state-of-the-art model for the next POI recommendation task.
% \end{itemize}
% \vspace{-0.3cm}
\subsection{Experimental Setup}
\label{sec:5_experimental}
% \subsubsection{Metrics}
% \label{sec:5_metrics}

\subsubsection{Baselines} We implement 8 baselines to evaluate our model.
\begin{itemize}[leftmargin=*]
\item \textbf{MC}~\cite{gambs2012next}is a conventional recommendation model that forecasts the subsequent item based on the probabilities derived from the Markov Chains.

\item \textbf{LSTM}~\cite{hochreiter1997long} is a variation of the RNN model designed to effectively model sequence through the simultaneous consideration of both long-term and short-term dependencies.

\item \textbf{ST-RNN}~\cite{liu2016predicting} extends the traditional RNN by incorporating the transformation matrix of temporal and spatial features.

\item \textbf{DeepMove}~\cite{feng2018deepmove} captures sequential patterns by integrating multiple attention mechanisms to model both the long-term and short-term interests of users.

\item \textbf{Flashback}~\cite{yang2020location} is an RNN-based model that places a strong emphasis on modeling historical check-in data.

\item \textbf{LSTPM}~\cite{sun2020go} utilizes a geo-nonlocal LSTM architecture to model the evolving patterns of long-term interests and a geo-dilated LSTM to capture the geographic dependencies of short-term interests. 

\item \textbf{STAN}~\cite{luo2021stan} employs an attention mechanism to learn the spatial-temporal relationships between non-adjacent POIs in the check-in sequence.

\item \textbf{GETNext}~\cite{yang2022getnext} is a Transformer-based model that utilizes all user mobility patterns to represent POIs.
GETNext is the state-of-the-art model for the next POI recommendation task.
\end{itemize}

% \textbf{(1) MC}~\cite{gambs2012next} is a conventional recommendation model that forecasts the subsequent item based on the probabilities derived from the Markov Chains. 
% \textbf{(2) LSTM}~\cite{hochreiter1997long} is a variation of the RNN model designed to effectively model sequence through the simultaneous consideration of both long-term and short-term dependencies.
% \textbf{(3) ST-RNN}~\cite{liu2016predicting} extends the traditional RNN by incorporating the transformation matrix of temporal and spatial features.
% \textbf{(4) DeepMove}~\cite{feng2018deepmove} captures sequential patterns by integrating multiple attention mechanisms to model both the long-term and short-term interests of users.
% \textbf{(5) Flashback}~\cite{yang2020location} is an RNN-based model that places a strong emphasis on modeling historical check-in data.
% \textbf{(6) LSTPM}~\cite{sun2020go} utilizes a geo-nonlocal LSTM architecture to model the evolving patterns of long-term interests and a geo-dilated LSTM to capture the geographic dependencies of short-term interests. 
% \textbf{(7) STAN}~\cite{luo2021stan} employs an attention mechanism to learn the spatial-temporal relationships between non-adjacent POIs in the check-in sequence.
% \textbf{(8) GETNext}~\cite{yang2022getnext} is a Transformer-based model that utilizes all user mobility patterns to represent POIs.
% GETNext is the SOTA model for our task.

\subsubsection{Metrics} We employ top-k Accuracy@1,5,10 to evaluate the recommendation performance of our model. 
Top-k accuracy measures the probability of having the correct label within the top-k predicted samples. 
To better understand the position of the correct label within the top-k predicted samples, we adopt NDCG@5,10 (Normalized Discounted Cumulative Gain) as an evaluation metric, which can reflect both the position and score of the correct label within the top-k predicted samples.
It should be noted that the value of NDCG@1 is equal to Acc@1, and we omit NDCG@1 in the results.
In addition, we use MRR (Mean Reciprocal Rank) to measure the average position of the correct label among all predicted samples. 
The higher value indicates better model performance.

\subsubsection{Settings} We employ AdamW as the optimizer, with the batch size being determined by the dataset size, and the learning rate is set to decay from 0.0002.
If the training error remains unchanged for a certain period, the training algorithm will terminate early, or it will stop after 200 epochs.
We evaluate the experiments on a GPU server, where the computing node is equipped with a 38-core Intel(R) Xeon(R) Platinum 8368 CPU @ 2.40GHz, 256GB RAM, and 1 NVIDIA A100 Tensor Core 40GB GPU.
% We evaluate the experiments on a multi-node GPU cluster, where each node is equipped with a 38-core Intel(R) Xeon(R) Platinum 8368 CPU @ 2.40GHz, 256GB RAM, and four NVIDIA A100 Tensor Core 40GB GPUs.
% Our experiment is conducted using PyTorch 1.7.1 and Python 3.7.
In our model, the hidden state size is set to 128, and the number of attention layers is set to 3. 
% Hyperparameter studies on these can be found in the Appendix.

% \vspace{-0.3cm}
\subsection{Results and Analysis}
\label{sec:5_results}
% Table~\ref{tab:re_gowalla}, Table~\ref{tab:re_tky} and Table~\ref{tab:re_toyota} illustrate the results of our proposed model compared to other baseline models on the three datasets.
Table~\ref{tab:result_acc} and Table~\ref{tab:result_ndcg} illustrate the results of our MobGT compared to other baseline models on the three datasets.
Overall, our model performs better on the Foursquare-TKY and Toyota datasets than on the Gowalla-NV dataset.
This can be attributed to the fact that the area of Nevada is significantly larger than that of Tokyo, which leads to more scattered and sparse user trajectories.
As shown in Table~\ref{tab:data-statstic}, the Gowalla dataset contains approximately 88k check-in records and 3,679 POIs, which are distributed over an area of 286,380 square kilometers in Nevada. 
The Toyota dataset contains approximately 540k check-in records and 8,011 POIs, which are distributed over the entire Tokyo metropolitan area of 2,194.07 square kilometers.
Due to the inherent sparsity of the datasets, the model's performance on Gowalla is markedly lower than that on Foursquare-TKY and Toyota.
Specifically, our model achieves an Acc@1 of 22.09\% on the Foursquare-TKY dataset, while on Gowalla, it can only reach 16.90\%.
% Although our model achieves an Acc@1 of 16.90\% on Gowalla, which is higher than the best baseline of 24.91\%, it is still considerably lower than the accuracy achieved on the other two datasets.

Nonetheless, MobGT outperforms the best baseline models on all datasets.
For instance, on the Toyota dataset, we achieve an Acc@1 accuracy of 26.34\%, whereas the state-of-the-art model LSTPM only achieves 19.18\%. 
% Our model also improves the NDCG@5 and NDCG@10 by 25.81\% and 22.28\%, respectively, compared to the best baseline model. 
Our model also improves the NDCG@5 by 25.81\% compared to the best baseline model. 
Similar results are observed on other datasets.
Furthermore, traditional Markov models are no longer able to learn complex transition states in trajectories, while sequence models such as LSTPM and STAN exhibit weaker overall recommendation performance than graph-based models such as GETNext. 
For example, on the Gowalla dataset, GETNext achieves an Acc@1 accuracy that is 53.23\% higher than that of LSTPM.

\begin{table*}[h]
\small
\renewcommand{\arraystretch}{0.92}
\caption{Performance comparison in Acc@k on three datasets of Gowalla, Foursquare, and Toyota. We highlight the best result in bold and underline the second-best result.}
\label{tab:result_acc}
\begin{tabular*}{17.5cm}{@{\extracolsep{\fill}}lccc|ccc|ccc}
\toprule
\multicolumn{4}{c|}{Gowalla-NV}                                                                                                                                                                                                                                                                & \multicolumn{3}{c|}{Foursquare-TKY}                                                                                                                                                                                                                  & \multicolumn{3}{c}{Toyota-TKY}    
\\
\midrule
                                     & Acc@1                                  & Acc@5                                                                  & Acc@10                                                                                                 & Acc@1                                  & Acc@5                                                                 & Acc@10                                                                                                   & Acc@1                                  & Acc@5                                                                 & Acc@10                                                                                                 \\
\midrule                                     
MC~\cite{gambs2012next}            & 0.0225                    & 0.0523                                        & 0.0668                                                                              & 0.0387          & 0.0769                  & 0.0925                                                 & 0.0190          & 0.0318                   & 0.0365                                                 \\
LSTM~\cite{hochreiter1997long}                                 & 0.0591                                 & 0.1396                                                                  & 0.1841                                                                                                 & 0.0944                                 & 0.2199                                                               & 0.2809                                                                                              & 0.1305                                 & 0.2309                                                                  & 0.2925                                                                                                \\
ST-RNN~\cite{liu2016predicting}                               & 0.1149                                 & 0.1277                                                                & 0.1362                                                                                                & 0.0245                                 & 0.0885                                                                 & 0.1131                                                                                              & 0.0221                                 & 0.0443                                                                 & 0.0634                                                                                               \\
DeepMove~\cite{feng2018deepmove}                             & 0.0732                                 & 0.1710                                                                  & 0.2265                                                                                                  & 0.1601                                 & 0.3299                                                                & 0.3996                                                                                               & 0.1614          & 0.2970                   & 0.3702                                             \\
Flashback~\cite{yang2020location}                            & 0.0703                                 & 0.1618                                                                & 0.2136                                                                                                & 0.1283                                 & 0.2867                                                               & 0.3506                                                                                                & 0.1853                                 & 0.3294                                                                & 0.3934                                                                                               \\
LSTPM~\cite{sun2020go}                                & 0.0883                                 & 0.1971                                                                & 0.2617                                                                                            & 0.1806                                 & 0.3844                                                                & 0.4659                                                                                                 & \underline{0.1918}          & \underline{0.3584}                   & \underline{0.4409}                                                 \\
STAN~\cite{luo2021stan}                                 & 0.0746                                 & 0.2005                                                                  & 0.2508                                                                                             & 0.1156          & 0.3063                  & 0.4110                     & 0.0800          & 0.2320                  & 0.3200                    \\
GETNext~\cite{yang2022getnext}                              & \underline{0.1353}                    & \underline{0.2516}                                         & \underline{0.2896}                                                     & \underline{0.2017}          & \underline{0.4173}                    & \underline{0.4940}                                           & 0.1062                                 & 0.2093                                                               & 0.2581                                                                                             \\
\midrule
\textbf{MobGT (Ours)} & \textbf{0.1690}           & \textbf{0.2733}                     & \textbf{0.3112}                         & \textbf{0.2209} &   \textbf{0.4298}  & \textbf{0.4945}   & \textbf{0.2634} & \textbf{0.4297}  & \textbf{0.4999}  \\
\bottomrule 
\end{tabular*}
\end{table*}

\begin{table*}[h]
\small
\renewcommand{\arraystretch}{0.92}
\caption{Performance comparison in NDCG@k and MRR on three datasets of Gowalla, Foursquare, and Toyota. We highlight the best result in bold and underline the second-best result.}
% \small{
\label{tab:result_ndcg}
% \tabcolsep=1cm
% \renewcommand\arraystretch{1}
% \resizebox{0.8\textwidth}{!}{
\begin{tabular*}{17.5cm}{@{\extracolsep{\fill}}lccc|ccc|ccc}
\toprule
\multicolumn{4}{c|}{Gowalla-NV}                                                                                                                                                                                                                                                                & \multicolumn{3}{c|}{Foursquare-TKY}                                                                                                                                                                                                                  & \multicolumn{3}{c}{Toyota-TKY}    
\\
\midrule
                                     & NDCG@5                                  & NDCG@10                                                                  & MRR                                                                                                 & NDCG@5                                  & NDCG@10                                                                 & MRR                                                                                                   & NDCG@5                                  & NDCG@10                                                                 & MRR                                                                                                 \\
\midrule                                     
MC~\cite{gambs2012next}             & 0.0389                    & 0.0437                                        & 0.0399                                                                              & 0.0602          & 0.0653                  & 0.0316                                                 & 0.0259          & 0.0274                   & 0.0475                                                 \\
LSTM~\cite{hochreiter1997long}                                 & 0.1007                                 & 0.1151                                                                  & 0.1031                                                                                                 & 0.1598                                 & 0.1796                                                               & 0.1567                                                                                              & 0.1861                                 & 0.2040                                                                  & 0.1864                                                                                                \\
ST-RNN~\cite{liu2016predicting}                               & 0.1206                                 & 0.1232                                                                & 0.1226                                                                                                & 0.0577                                 & 0.0655                                                                 & 0.0552                                                                                              & 0.0340                                 & 0.0403                                                                 & 0.0370                                                                                               \\
DeepMove~\cite{feng2018deepmove}                               & 0.1240                                 & 0.1419                                                                  & 0.1255                                                                                                  & 0.2494                                 & 0.2721                                                                & 0.2405                                                                                               & 0.2319          & 0.2556                   & 0.2311                                             \\
Flashback~\cite{yang2020location}                            & 0.1175                                 & 0.1343                                                                & 0.1203                                                                                               & 0.2116                                 & 0.2324                                                               & 0.2033                                                                                                & 0.2616                                 & 0.2822                                                                & 0.2574                                                                                               \\
LSTPM~\cite{sun2020go}                                 & 0.1441                                 & 0.1648                                                                & 0.1456                                                                                            & 0.2877                                 & 0.3142                                                                & 0.2758                                                                                                 & \underline{0.2790}          & \underline{0.3056}                   & \underline{0.2752}                                                 \\
STAN~\cite{luo2021stan}                                 & 0.1388                                 & 0.1549                                                                  & 0.1321                                                                                             & 0.2116          & 0.2452                  & 0.2039                     & 0.1557          & 0.1838                  & 0.1548                    \\
GETNext~\cite{yang2022getnext}                              & \underline{0.1967}                    & \underline{0.2084}                                         & \underline{0.1908}                                                     & \underline{0.3160}          & \underline{0.3408}                    & \underline{0.3017}                                           & 0.1600                                 & 0.1759                                                               & 0.1586                                                                                             \\
\midrule
\textbf{MobGT (Ours)} & \textbf{0.2234}           & \textbf{0.2356}                     & \textbf{0.2174}                         & \textbf{0.3344} &   \textbf{0.3544}  & \textbf{0.3165}   & \textbf{0.3510} & \textbf{0.3737}  & \textbf{0.3437}  \\
\bottomrule 
\end{tabular*}
% }
% \vspace{-0.5cm}
\end{table*}

\begin{table*}[!tpb]
\caption{The performance comparison among the full model and different variants without part of the component. We highlight the best result in bold.}
\small
\renewcommand{\arraystretch}{0.92}
% \centering
\label{tab:ablation}
% \resizebox{0.8\textwidth}{!}{
\begin{tabular*}{17.5cm}{@{\extracolsep{\fill}}lcccccc}
\toprule
Model/Metric           & Acc@1           & Acc@5           & Acc@10                    & NDCG@5          & NDCG@10         & MRR             \\ \midrule 
\textbf{MobGT (complete)}              & \textbf{0.1690} & \textbf{0.2733} & \textbf{0.3112}  & \textbf{0.2234} & \textbf{0.2356} & \textbf{0.2174} \\ 
w/o Spatial Graph        & 0.1675          & 0.2609          & 0.2980                    & 0.2148          & 0.2251          & 0.2086          \\
w/o Temporal Graph        & 0.1606          & 0.2583          & 0.2928                    & 0.2090          & 0.2217          & 0.2054          \\
w/o Global Graph        & 0.1580          & 0.2596          & 0.2975                    & 0.2128          & 0.2250          & 0.2079          \\
w/o ST Graph Attention & 0.1590          & 0.2575          & 0.2986                    & 0.2124          & 0.2256          & 0.2086          \\
w/o Context         & 0.1116          & 0.1327          & 0.1627                    & 0.1221          & 0.1315          & 0.1286          \\
w/o Tail Loss   & 0.1588          & 0.2665          & 0.3017                   & 0.2161          & 0.2275          & 0.2102  \\
w/o Category \& Tail Loss   &0.1480 &0.2396 &0.2801  &0.1978 &0.2110 &0.1956 \\
\bottomrule      
\end{tabular*}
% }
% \vspace{-0.5cm}
\end{table*}

\subsection{Ablation Study} We conduct rigorous ablation experiments to demonstrate the impact of each component in our proposed model on the overall prediction performance of the Gowalla dataset. The results are presented in Table~\ref{tab:ablation}. 
Specifically, we perform 8 experiments: (1) the full MobGT, (2)without Spatial global graph, (3)without Temporal global graph, (4) without all global graphs (Spatial, Temporal, and Category), (5) without ST Graph Attention component, (6) without related contextual information including user, category, frequency and check-in time, only using POI embedding from the global view, (7) without proposed Tail Loss $\mathcal{L}_{tail}$ and utilizing cross-entropy loss as the category prediction loss, (8) without category decoder and utilizing Tail Loss $\mathcal{L}_{tail}$ as the POI prediction loss.
% In these experiments, all hyperparameter settings are kept consistent, with only one component being altered each time, and the results are presented in Table~\ref{tab:ablation}.
The complete MobGT evidently achieved the best performance, while for the remaining components, the results indicate that the ST contextual component has the most significant influence on the overall model performance. 
For instance, when not utilizing the spatial-temporal context to enrich the POI embeddings, Acc@1 decreases by 33.96\% compared to the complete model. 
Other components also contribute to the overall enhancement of the model, such as the global view module component, which leads to a 6.96\% improvement in Acc@1.

\subsection{Hyperparameter Study}
\begin{figure}[!ht]
	\centering	
	\subfigure[Hidden Dimension]{
		\includegraphics[width=0.22\textwidth]{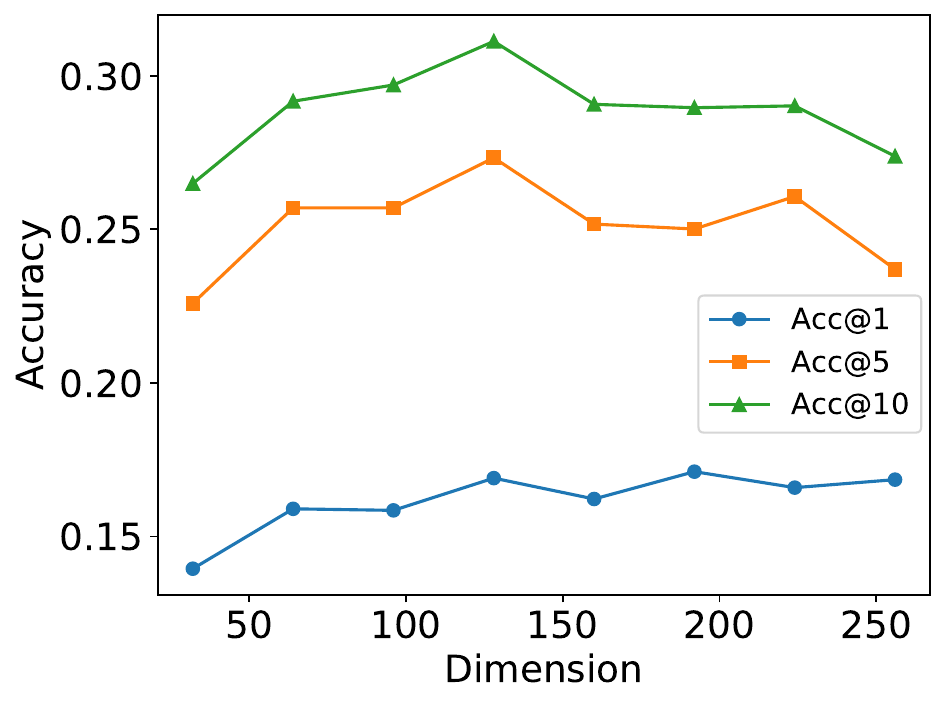}
		\label{fig:hyper_1}
	}\ \ \
	\subfigure[Attention Layer]{
		\includegraphics[width=0.22\textwidth]{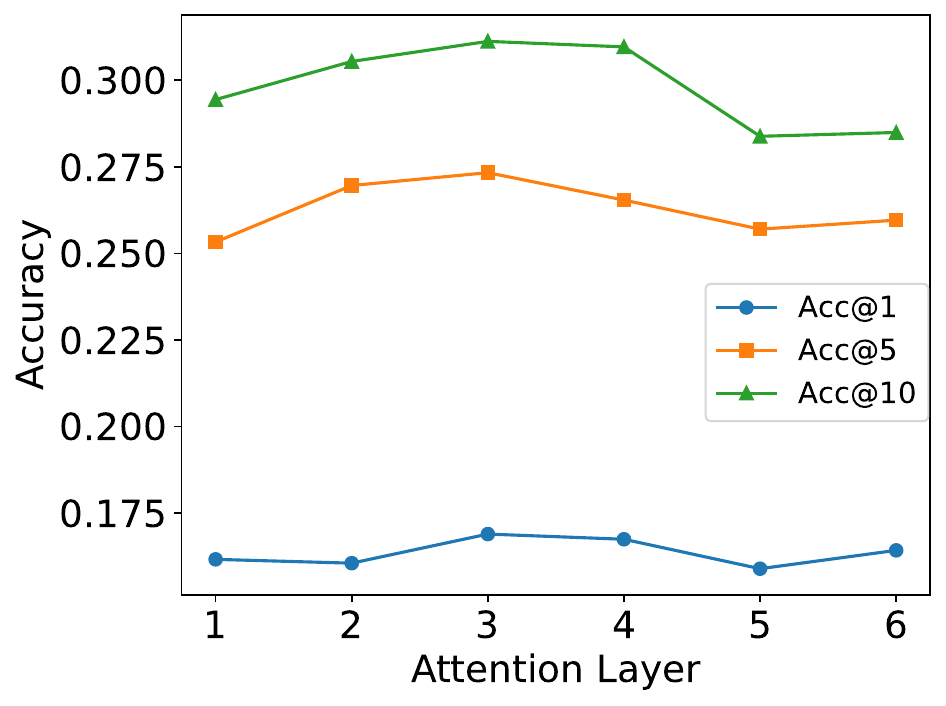}
		\label{fig:hyper_2}
	}
	\caption{Impact of hidden dimension $d$ and attention layer $l$ on Gowalla dataset.}\label{fig:hyper}
 \vspace{-0.5cm}
\end{figure}

To validate the stability of our model, we conduct corresponding hyperparameter experiments. 
We set the dimension of the hidden layers $d$ in the model to an arithmetic sequence ranging from 32 to 256, successively testing the impact of different hidden layer dimensions on the model. 
As illustrated in Fig.~\ref{fig:hyper_1}, on the Gowalla dataset, when $d=128$, although the performance in terms of Acc@1 is slightly lower than when $d=192$, the remaining metrics are significantly higher than other settings. 
Overall, our model's prediction performance is not sensitive to the $d$ hyperparameter. 
Excluding the settings, $d=32$ and $d=256$, the variation rate for Acc@1 is less than 8\%, and the variation rate for MRR is less than 5\%, indicating a relatively insignificant impact.
On the other hand, we also test the number of attention layers $l$ in the model. 
As shown in Fig.~\ref{fig:hyper_2}, we draw a similar conclusion: although the number of attention layers changes continuously, the variation rate for Acc@1 is less than 6\%, and the variation rate for MRR is less than 7\%. 
The model achieves the best performance when we set $l=3$.

\subsection{Case Study} 
\begin{figure}[!ht]
	\centering	
	\subfigure[Zoom In View]{
		\includegraphics[width=0.22\textwidth]{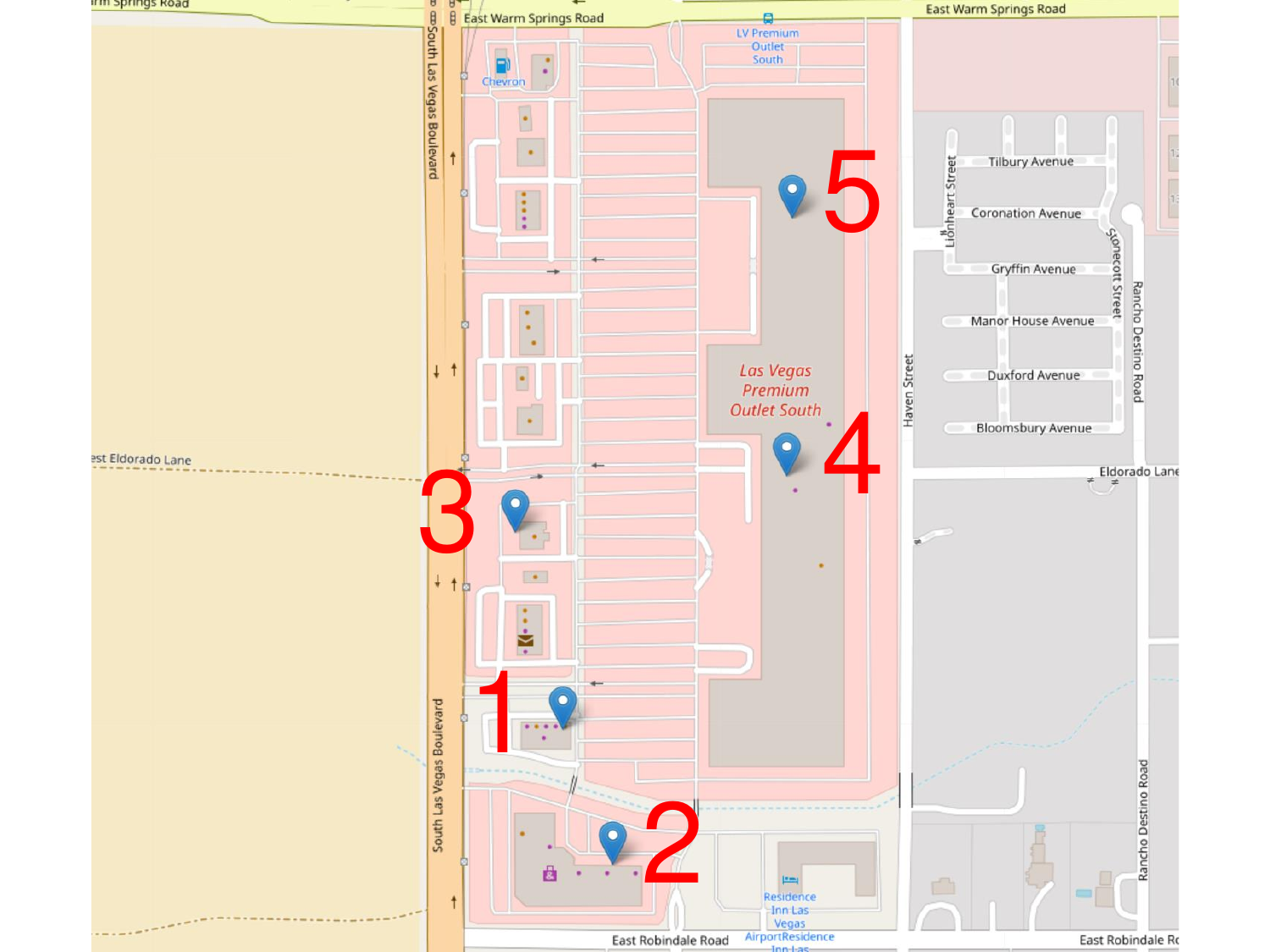}
		\label{fig:data_interpret_1}
	}\ \ \
	\subfigure[Zoom Out View]{
		\includegraphics[width=0.22\textwidth]{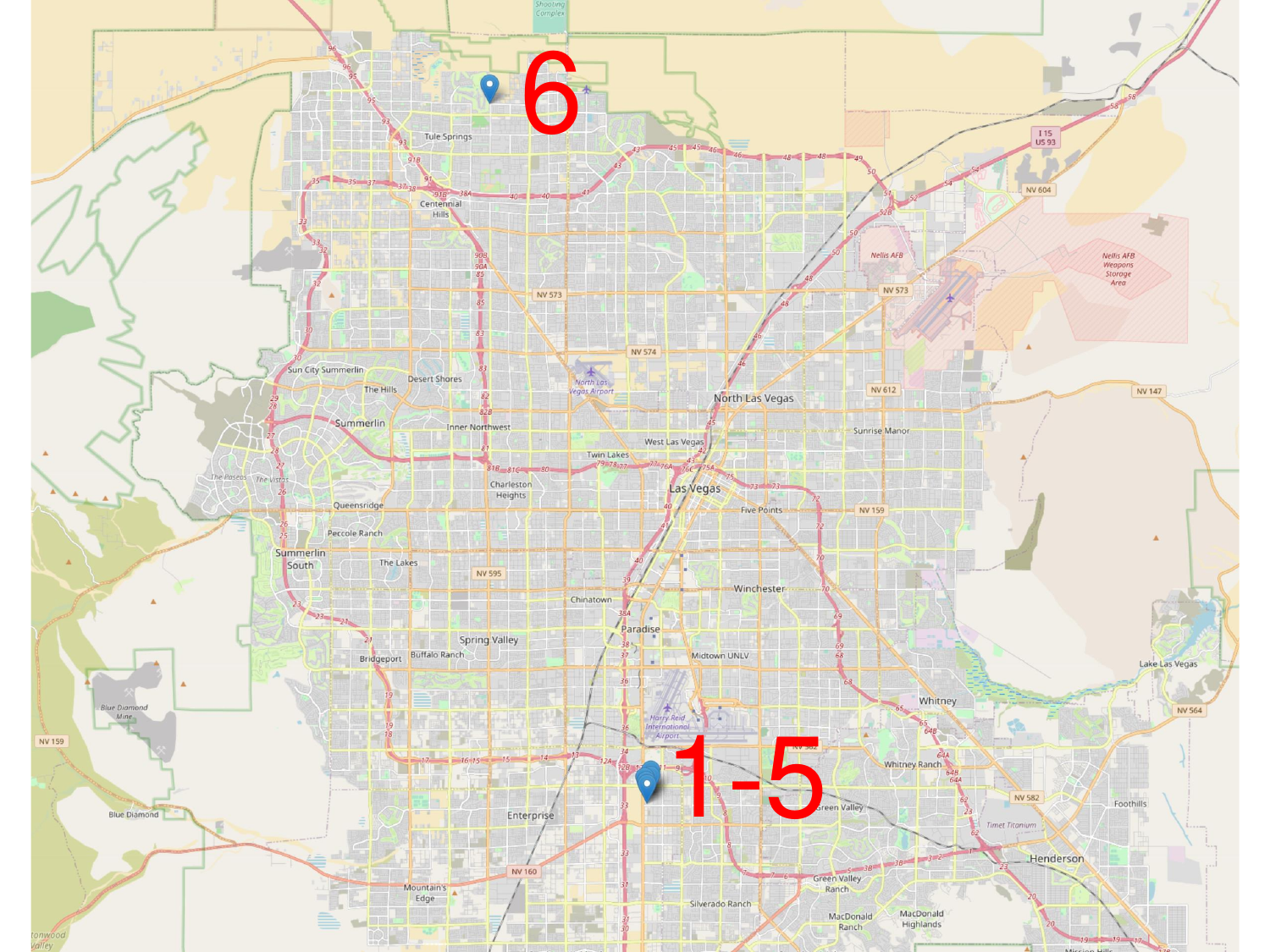}
		\label{fig:data_interpret_2}
        }
        \subfigure[Correlation Matrix]{
		\includegraphics[width=0.3\textwidth]{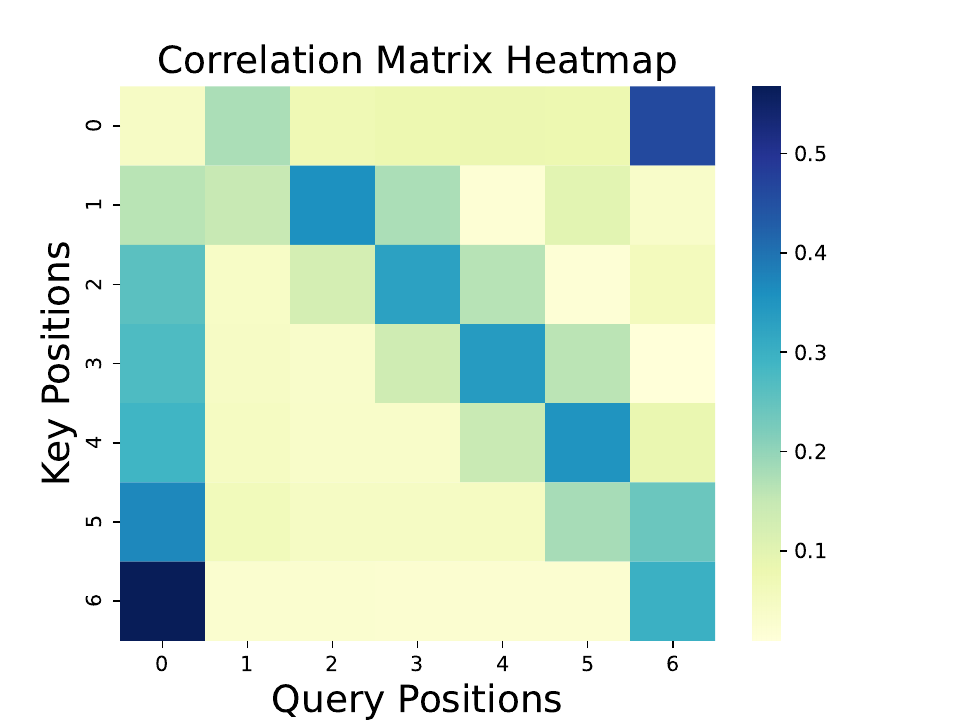}
		\label{fig:data_interpret_3}
	}
	\caption{A visualization sample from the Gowalla dataset. We note query and key weights in our ST Graph Attention as a correlation matrix. It should be noted that because POI 6 is far from POI 1-5, we offer two different views (zoom in and out).}\label{fig:data_interpret}
\vspace{-0.3cm}
\end{figure}

The MobGT model possesses good interpretability, such as predicting the next POI based on a given Traj sample containing POI 1-6. 
From Fig.~\ref{fig:data_interpret_1}, we can observe that the user frequently checks in at a place similar to a shopping mall, mainly for shopping or dining. 
However, based on Fig.~\ref{fig:data_interpret_2}, POI 6 is very far from POI 1-5, and it is highly likely that the user checked in some POIs in a shopping mall and then drove back home. 
We can also observe from Fig.~\ref{fig:data_interpret_3} that Center Node 0 has the strongest association with POI 6, and POI 6 also has the strongest association with Center Node 0. 
Therefore, the final prediction result is likely to be POI 6 (according to the label, the next visited POI is still POI 6). 
For the case where the user checks in twice at the same location, we consider that this is due to data loss on the Gowalla platform or the user's behavior of checking in again when leaving home.
% \vspace{-0.2cm}

% \vspace{-0.5cm}

\section{Conclusion}
\label{sec:6_conclusion}
This paper proposes a novel next POI recommendation model MobGT based on graph and spatial-temporal attention by exploring transition patterns of trajectories in all users and each individual user. 
Specifically, our model considers the mobility preferences of all users and the inherent location information between POIs as components of overall POI transition behavior. 
We use GNN to learn this POI preference, which is then used for each user's unique mobility subgraph. 
The spatial-temporal context information of all POIs is encoded using proposed ST attention combined with the degree and structure information unique to graph structures.
Finally, we also proposed Tail Loss to address the long tail problem inherent in these POI check-in datasets
Extensive experiments on three real-world datasets demonstrate that our proposed model outperforms state-of-the-art models in all metrics. 
We also validate the stability of our model under different parameters and provide visualization results to demonstrate the interpretability of our model in predicting the next POI. 
For future work, we plan to build a module to specifically learn the features of long-tail data to further enhance the interpretability of our model in solving long-tail problems.
Another possible direction is to improve the representation of user embedding, such as performing clustering operations based on user movement trajectories in advance, so that the model can more easily learn POI representation from similar user trajectories.

\begin{acks}
This work was supported by JST SPRING (JPMJSP2108), JSPS KAKENHI Grant Numbers JP21K17749 and JP21K21280, and Initiative on Promotion of Supercomputing for Young or Women Researchers, Information Technology Center, The University of Tokyo.
\end{acks}

\bibliographystyle{ACM-Reference-Format}
\bibliography{reference}


\end{document}